\documentclass[preprint,3p,times,onecolumn]{elsarticle}

\usepackage[caption=false,font=footnotesize]{subfig}

\usepackage{amssymb, amsmath, amsthm, bm}

\usepackage{pgfplots}
\pgfplotsset{compat=newest}
\pgfplotsset{
tick label style={font=\small},
label style={font=\small},
legend style={font=\footnotesize, legend cell align=left, fill=none, draw=none},
major grid style={line width=0.2pt},
}
\usetikzlibrary{plotmarks}
\usetikzlibrary{arrows.meta}
\usepgfplotslibrary{patchplots}
\usepackage{color}

\usepackage{graphicx}
\graphicspath{{figures/}}

\usepackage{xurl}

\usepackage[mathlines]{lineno}

\usepackage{multirow}

\newtheorem{Property}{Property}
\newtheorem{Fact}{Fact}

\usepackage{setspace}

\newlength\twofigwidth
\begin{document}

\begin{frontmatter}

\title{Recovering Sign Bits of DCT Coefficients\protect\\in Digital Images as an Optimization Problem}

\author[usa]{Ruiyuan~Lin}

\author[cn-hnu]{Sheng~Liu}

\author[cn-pku]{Jun~Jiang}

\author[uk]{Shujun~Li\corref{corr}}

\author[cn-xtu]{Chengqing Li\corref{corr}}

\author[usa]{C.-C.~Jay~Kuo}

\address[usa]{Multimedia Communication Lab, Department of Electrical and Computer Engineering, University of Southern California, US}

\address[cn-hnu]{School of Computer Science and Electronic Engineering, Hunan University, Changsha 410082, Hunan, China}

\address[cn-pku]{School of Software \textup{\&} Microelectronics, Electronic \textup{\&} Information Engineering, Peking University, China}

\address[uk]{Institute of Cyber Security for Society (iCSS) \textup{\&} School of Computing, University of Kent, UK}

\address[cn-xtu]{School of Computer Science, Xiangtan University, Xiangtan 411105, Hunan, China}

\cortext[corr]{Corresponding author.}

\begin{abstract}
Recovering unknown, missing, damaged, distorted, or lost information in DCT coefficients is a common task in multiple applications of digital image processing, including image compression, selective image encryption, and image communication. This paper investigates the recovery of sign bits in DCT coefficients of digital images, by proposing two different approximation methods to solve a mixed integer linear programming (MILP) problem, which is NP-hard in general. One method is a relaxation of the MILP problem to a linear programming (LP) problem, and the other splits the original MILP problem into some smaller MILP problems and an LP problem. We considered how the proposed methods can be applied to JPEG-encoded images and conducted extensive experiments to validate their performances. The experimental results showed that the proposed methods outperformed other existing methods by a substantial margin, both according to objective quality metrics and our subjective evaluation.
\end{abstract}
\begin{keyword}
Discrete cosine transform (DCT); sign bit recovery; optimization; integer programming; linear programming; selective encryption
\end{keyword}
\end{frontmatter}

\section{Introduction}

The discrete cosine transform (DCT) was first proposed by Nasir Ahmed in 1972~\cite{Ahmed1991DCT}. As a signal analysis tool, DCT can be used to convert a discrete input signal with $n$ samples into $n$ frequency components, which are also called DCT coefficients. The first of all DCT coefficients is normally called the DC (direct current) coefficient, representing the zero-frequency component or the average amplitude of the input signal. Other DCT coefficients all represent the amplitude of a specific non-zero frequency component and are normally called AC (alternate current) coefficients. Although DCT is not the optimal signal decorrelation transform, it can approximate the optimal solution well under a wide range of conditions and can be very easily implemented, therefore it has been widely utilized in many signal processing applications~\cite{Langelaar:DEW:TIP2001, Das:DEW:TSP2005, Wu:DAC:TMM16}. Especially, 2-D blockwise DCT has been widely adopted in image and video compression standards including JPEG~\cite{Wallace:JPEG:CACM1991}, MPEG-1/2, MPEG-4, AVC/H.264~\cite{Jack:video:2011}, and HEVC~\cite{Sullivan:HEVC:TCSVT2012}. Meanwhile, researchers have extensively investigated characteristics of DCT in different applications~\cite{Lakhani:DDCT:TCSVT2000, Lam:DCT:TIP2000, jiang2002spatial:TSP02, Xu:MCR:GLOBECOM2011, Hung:neural:VCIP18, Sean:graph:TIP19}.

Sensitive multimedia data should be encrypted before transmission to prevent unauthorized access and protect individual privacy. Due to the large volume of multimedia data and some other needs, such as format-compliance and perceptual encryption~\cite{Li:PEIV:Bookchapter2012}, selectively encrypting a small part of important data becomes a natural choice among researchers~\cite{Zeng:DCTENC:TMM2003, Li:MPEG:CSVT07, Dufaux:Scrambling:TCSVT2008, Sohn:encrypt:TCSVT11}. However, some special properties of multimedia data, such as the strong correlation of neighboring pixels, and the more stable distribution of DCT coefficients, make many proposed selective encryption schemes of digital images and videos insecure against various attacking methods in different settings~\cite{Li:PEIV:Bookchapter2012}. Among all the attacks, one group works under the ciphertext-only condition and tries to recover an encrypted image/video to reveal more visual information than simple error-concealment attacks can recover, where ``error-concealment attacks" refer to attacks that simply replace encrypted information with some simple values (e.g., zeros). For instance, in~\cite{Uehara:DCRecover:TIP06}, Uehara {\em et al.} demonstrated that encrypted DC coefficients can be approximately restored from known AC ones by exploiting the strong correlation between adjacent pixels, resulting in recovered images with satisfactory visual quality. This work was further improved by Li et {\em et al.} by using an optimization-based approach to minimize the so-called under/over-flow rate of pixel values caused by error propagation~\cite{Li:DCT:ICIP2010}. Soon after, Li {\em et al.} proposed a more general method for restoring an arbitrary set of missing DCT coefficients from known ones, and formulated the problem as a linear programming (LP) problem that aims to minimize the sum of absolute differences between neighboring pixels, which was shown to outperform all previously proposed methods significantly~\cite{Li:DCT:ICIP2011}. Note that the DCT coefficient recovery method can be used not only for attacking selective encryption schemes but also in other less security-related applications, such as image compression, and image recovery.

As the most significant bit of a DCT coefficient, the sign bit plays an important role in DCT-based image and video coding and encryption. In many image and video coding standards, sign bits are separately encoded. Researchers have also looked at how to more efficiently encode sign bits to reduce informational redundancies in the encoded image and video bitstream~\cite{Ponomarenko:sign:IPAS07, Koyama:sign:PCS12, Lakhani:sign:TIP13, Filippov:sign:SIP19}. Recently, Suzuki {\em et al.} predicted the sign bits of DCT coefficients only from their magnitudes by using a DNN to solve a convex optimization and encoded the difference between the results and the original signs after the quantization process in the standard JPEG pipeline, where the DNN was trained with numerous images \cite{suzuki2022compressing}. However, the average sign recovery accuracy is about 65\%-70\%, so the method is quite limited in the scenario that partial signs are lost.
When the ratio of unknown signs is less than the least recovery accuracy,
the method would incur even more sign loss. 

In the context of image and video encryption, randomly flipping sign bits has been widely adopted as a general component in selective encryption schemes due to its capability to maintain format compliance and also size-preservation while being very easy to implement and highly efficient~\cite{Shi:encrypt:ICM98, Dufaux:Scrambling:TCSVT2008, Sohn:encrypt:TCSVT11, Wang:encrypt:TCSVT13, Hofbauer:encrypt:ICASSP14, Ma:AVC:TETC16, Peng:HEVC:TCSVT20}. Despite the wide use of sign bits, we rarely see research investigating advanced methods for recovering unknown sign bits of DCT coefficients, where advanced methods refer to those that are not based on simple error-concealment strategies. This gap leads to a lack of understanding of the security of selective image and video encryption schemes based on sign bit encryption and also on more efficient sign bit prediction and compression methods for image and video coding.

In this paper, we fill the above-mentioned research gap by formulating the sign bit recovery problem as an optimization problem that aims to minimize the sum of absolute differences between neighboring pixels, following a similar vein of the DCT coefficient recovery model reported in~\cite{Li:DCT:ICIP2011}. Different from and more challenging than the work in~\cite{Li:DCT:ICIP2011}, a sign bit is a binary value, so the optimization model is a mixed integer linear programming (MILP) problem, which is NP-hard in general. To solve the problem efficiently, we propose two approximation methods that can obtain reasonably good visual quality with a manageable time complexity. In the first method, we relax the MILP problem to an LP problem by replacing the binary unknown variables with continuous values and then estimate the sign bits based on the solution of the relaxed LP problem. In the second method, we divide the image into sufficiently small sub-images as~\cite{zhou:CS:TM21}, then solve a MILP problem for each sub-image independently, and finally refine the merged result by solving a global LP or MILP optimization problem. We extended the proposed methods to handle JPEG images used in real-world applications so that the methods can take special encoding rules about sign bits and DCT coefficients. To demonstrate the performance of the proposed sign bit methods and how they work with different encoding parameters of the JPEG standard, we conducted extensive experiments with a set of 30 standard test images. We also compared the performance of the proposed methods with other native ones and a simplified version based on a relaxed LP model alone. The experimental results showed that the proposed methods were able to achieve satisfying visual quality with a practical time complexity, and remarkably outperformed other baseline methods. Our work can not only provide more insights on designing and evaluating selective multimedia encryption schemes but also guidance on how to design more efficient image and video coding methods and how to recover damaged, distorted, or lost sign bits in error-prone environments.

The rest of the paper is organized as follows. Section~\ref{sec:review} reviews the related work. Section~\ref{sec:proposed} explains our two proposed sign bit recovery methods with details. Experimental results are given in Section~\ref{sec:results}. The last section concludes the paper.

\section{Related Work}
\label{sec:review}

This section is organized to show closely related work in four areas: image and video encryption schemes using sign bit encryption; simple (not optimization-based) methods for recovering DCT coefficients; optimization-based methods for recovering DCT coefficients; and known methods for recovering sign bits of DCT coefficients.\footnote{This paper focuses more on recovery problems of DCT coefficients, so does not cover related work on other transforms, e.g., work in \cite{Pei:DFT:TSP2022} on recovering binary signals from a limited number of DFT coefficients.}

\subsection{Sign bit encryption for image and video encryption and privacy protection}

As mentioned in Section~1, randomly flipping sign bits has been widely used for image and video encryption~\cite{Shi:encrypt:ICM98, Wang:encrypt:TCSVT13, Hofbauer:encrypt:ICASSP14, Peng:HEVC:TCSVT20}. Here, we briefly introduce some representative work. Wang {\em et al.} designed a tunable encryption scheme for H.264/AVC videos, in which the encryption strength is adjusted by selecting some encryption objects: intra prediction modes, sign bits of non-zero coefficients and sign bits of motion vectors~\cite{Wang:encrypt:TCSVT13}. Hofbauer {\em et al.} proposed an encryption scheme for HEVC videos that solely encrypts some sign bits of AC coefficients in the encoded bitstream to distort the visual information strongly while maintaining full format-compliance and size-preservation~\cite{Hofbauer:encrypt:ICASSP14}, but also acknowledged that encrypting AC coefficients' sign bits alone cannot guarantee full confidentiality based on some analysis~\cite{Hofbauer:AC:ICIP2015}. Hofbauer {\em et al.}, however, did not propose a method to recover the encrypted sign bits to improve the visual quality of recovered images. Some researchers also proposed to use sign bit encryption for video surveillance systems for privacy protection purposes~\cite{Dufaux:Scrambling:TCSVT2008, Sohn:encrypt:TCSVT11, Ma:AVC:TETC16}. For instance, Dufaux {\em et al.} designed a privacy protection algorithm that encrypts only privacy-sensitive regions through randomly flipping sign bit while keeping the surveilled scenes comprehensible~\cite{Dufaux:Scrambling:TCSVT2008}.

\subsection{Simple image and video recovery against transmission errors and encryption}

If errors occur during the transmission and storage of image data, there may exist some incomprehensible or simply blank areas in the decoded version \cite{cqli:CS:TMM23}. To address this problem, some error-concealment techniques were proposed to recover such corrupted areas via exploiting the strong correlation between the target area and its surrounding areas~\cite{Natale:RecoverDCT:JSAC2000, Park:DCT:ISCAS02, Bingabr:DCT:CASVT2004}. Bingabr and Varshney designed a recovery algorithm that can accurately correct corrupted coefficients in one DCT block with the assistance of reference information accurately received from an extra channel~\cite{Bingabr:DCT:CASVT2004}. Some researchers also proposed to use supervised machine learning methods to recover corrupted DCT coefficients, e.g., using a trained neural network~\cite{Natale:RecoverDCT:JSAC2000}. Park {\em et al.} proposed an estimation algorithm for corrupted DCT coefficients based on projections onto convex sets, in which the surrounding undamaged blocks are extracted to form a convex hull for reference~\cite{Park:DCT:ISCAS02}. For errors occurring in video data, in addition to spatial information, the temporal redundancy between frames also can be exploited to recover corrupted DCT coefficients~\cite{Park:DCT:TCSVT1997, Tsekeridou:DCT:TCSVT2000}. For example, Tsekeridou {\em et al.} devised an error-concealment method for MPEG-2 videos based on spatio-temporal video redundancy and block-matching principles~\cite{Tsekeridou:DCT:TCSVT2000}.

In multimedia selective encryption, there are always components that are excluded from encryption to avoid encrypting the whole bitstream. In this case, such unencrypted data can be potentially used to estimate encrypted ones~\cite{Li:PEIV:Bookchapter2012, Li:MPEG:CSVT07}. Some sketch attacking methods were proposed to obtain an estimated image of low visual quality~\cite{Li:JPEG:IJCM2007, Minemura:sketch:ICIP2012, Minemura:Sketch:ASIPA2014, Minemura:sketch:TCSVT2017}. In \cite{Li:JPEG:IJCM2007}, Li and Yuan highlighted that the number of zero coefficients in each block is closely related to image texture and edge information, which generally remains unchanged to preserve the compression ratio in existing JPEG encryption algorithms. Taking advantage of this observation, they proposed a nonzero count attack on encrypted JPEG images, generating a rough sketch. Following a similar idea, Minemura {\em et al.} proposed three improved sketch attacks that do not require manual adjustment of thresholds and can generate images of higher visual quality~\cite{Minemura:sketch:ICIP2012}. Subsequently, these attack methods were extended to attack encryption schemes for H.264/AVC videos~\cite{Minemura:Sketch:ASIPA2014, Minemura:sketch:TCSVT2017}.

\subsection{Optimization-based DCT coefficient recovery}

Some DCT coefficients might be absent due to selective encryption, transmission errors, multimedia coding, or malicious removal. When only DC coefficients are missing, they can be estimated sequentially from the known AC ones by exploiting the strong correlation in multimedia data~\cite{Uehara:DCRecover:TIP06, Li:DCT:ICIP2010, Li:DCT:ICIP2011}. This is the so-called DC recovery problem. In~\cite{Uehara:DCRecover:TIP06}, Uehara {\em et al.} summarized two properties of most natural images, which were the cornerstone of their recovery method. They estimated the missing DC coefficients one by one by minimizing the sum of the absolute difference of boundary pixels between the current DCT block and its neighboring ones. Then, they adjusted the estimated DC coefficients to keep pixel values within the valid range. The final output image was the average of four images obtained by four different scanning directions. Li {\em et al.} pointed out that there was a serious error propagation effect in the estimation process of Uehara {\em et al.}'s method, and suggested immediately making adjustments after estimating each DC coefficient to alleviate such error propagation~\cite{Li:DCT:ICIP2010}. Moreover, they proposed to minimize the so-called under/over-flow rate of pixel values to improve the visual quality of recovered images. Qiu {\em et al.} designed a similar method to improve the error resistance for JPEG image transmission~\cite{Qiuhan:DC:2019}.

Generalizing the DC recovery problem, Li {\em et al.} defined a more general problem of recovering an arbitrary set of missing DCT coefficients from other known ones, which was formed as an optimization problem that can be solved as an LP problem~\cite{Li:DCT:ICIP2011}. The optimization aims to minimize the sum of the absolute differences of all neighboring pixels of the recovered image. In addition to being much more general, the new general method can also significantly surpass previous methods on the DC recovery problem. Later, a partition strategy~\cite{Li:DCT:SPIC2017} was introduced to reduce the time complexity of solving the optimization problem. The image is divided into multiple groups by image segmentation, each of which is separately recovered using LP, and then the brightness of each group is adjusted to minimize the discontinuing artifacts. As the special case of DCT recovery, the DC recovery problem can be modeled as the dual of a min-cost flow problem and then be solved with the corresponding algorithm, where the time complexity is drastically reduced from $O(n^2)$ to $O(n^{3/2})$~\cite{Li:LevelingGrid:ALENEX2012}.

The LP model in~\cite{Li:DCT:ICIP2011} was also extended to recover undecoded coefficients in distributed video coding scheme~\cite{Ali:LPA:DICTA2013}, in which an additional temporal smoothness maximization is introduced into the optimization process. In addition, Wang {\em et al.} modified the objective function of the optimization model in~\cite{Li:DCT:ICIP2011} and added a regularization term to restore part of the image from the structured side information~\cite{Li:Restore:SPIC2014}. In~\cite{Chen:HEVC:CASVT18}, Chen {\em et al.} proposed an optimization method to estimate DC coefficients for further compression, which exploits directional texture information of neighboring blocks and solves an optimal offset in a closed form.

\subsection{Sign bit recovery of DCT coefficients}

So far, sign bit recovery mainly exists in the decompression of image and video data~\cite{Ponomarenko:sign:IPAS07, Koyama:sign:PCS12, Rad:sign:ICIP13, Lakhani:sign:TIP13, Filippov:sign:SIP19, Tsutake:sign:ICIP21, Sullivan:HEVC:TCSVT2012}. In image and video compression standards based on the blockwise DCT, the redundancy among pixels within each DCT block is exploited to a large extent, while the redundancy between blocks especially between non-neighboring blocks is mostly unexplored. More specifically, the high correlation among pixels at the boundary between adjacent blocks is under-utilized, which can be used to predict the signs of coefficients to further improve compression performance. The reason for such under-exploration is the need for the encoding process to be in real-time, so a more complicated global optimization process is often considered too heavy.

In~\cite{Ponomarenko:sign:IPAS07}, Ponomarenko {\em et al.} proposed a sign bit prediction algorithm for lossy image compression, in which some sign bits are selected for compression in such a way that the border pixels reconstructed from inverse DCT are closest to those estimated from the previously-decoded blocks in spatial domain. To eliminate the expensive computational costs of transforming between spatial and frequency domains, Rad {\em et al.} proposed a sign bit recovery method that can operate in the frequency domain solely~\cite{Rad:sign:ICIP13}. They estimated the sign bits of five low-frequency coefficients and categorized DCT blocks into five patterns, each of which is treated with a different predictor. In~\cite{Lakhani:sign:TIP13}, Lakhani integrated the proposed sign bit prediction algorithms for some significant coefficients into a modified JPEG codec. Sign bit data hiding technique was introduced in the HEVC standard, in which the sign bit of a non-zero coefficient is omitted under some conditions~\cite{Sullivan:HEVC:TCSVT2012}.

Although recovery of DCT coefficients has been actively researched, the possibility of recovering encrypted, unknown, missing, or damaged sign bits from a digital image has been much less studied in the literature. This paper enriches the research by proposing the first sign bit recovery methods by modeling the recovery problem as an optimization problem, which can achieve a good recovery performance while having a practically small computational complexity.

\section{Proposed Methods}
\label{sec:proposed}

In this section, we first introduce the primary optimization model for the sign bit recovery problem. Then, two approximation methods, one based on linear programming (LP) and the other on hierarchical mixed integer linear programming (MILP), are presented to solve the problem efficiently.

\subsection{The model}

To facilitate the description and establishment of the optimization model, we first present two properties of pixel values of digital images, first summarized in~\cite{Uehara:DCRecover:TIP06}.

\begin{Property}
\label{prop:neighbor}
The difference between any two neighboring pixel values of a natural image is a Laplacian variate with a zero mean and a small variance.
\end{Property}

\begin{Property}
\label{prop:range}
For each block, pixel values of AC coefficients constrain the value of its DC coefficient.
\end{Property}

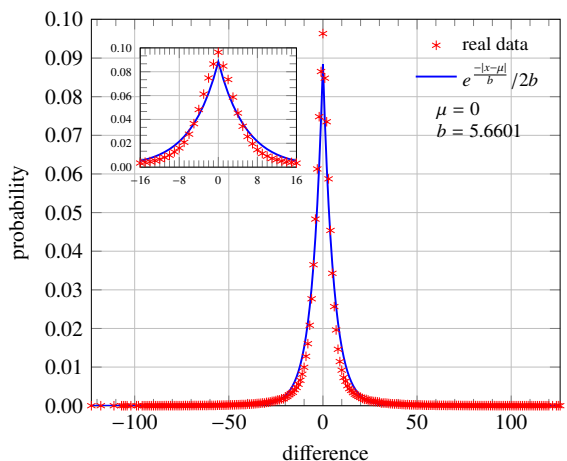
\begin{figure}[!htb]
\centering
\begin{tikzpicture}[scale=0.9]
\begin{axis}[
  xlabel={difference},
  ylabel={probability},
  xmin=-123, xmax=126,
  minor x tick num=4,
  ymin=0, ymax=0.1,
  ytick={0, 0.01, ..., 0.1},
  scaled y ticks=false,
  yticklabel style={
    /pgf/number format/precision=2,
    /pgf/number format/fixed,
    /pgf/number format/fixed zerofill,
  },
  xmajorgrids, ymajorgrids,
]
\addplot [
  red,
  only marks,
  mark=asterisk,
] table {figures/neighbour_distribution.dat};
\addlegendentry{real data};

\newcommand\B{5.6601}
\addplot [
  domain=-123:126,
  samples=250,
  blue,
  line width=0.8pt,
] {exp(-abs(x)/\B) / (2*\B)};
\addlegendentry{$e^{\frac{-|x-\mu|}{b}}/2b$};

\node[right, align=left, font=\footnotesize]
at (axis cs:56,0.074) {$\mu=0$\\ $b=\B$};
\end{axis}

\begin{axis}[
  width=110,
  height=95,
  at={(10,200)},
  xmin=-16, xmax=16,
  minor x tick num=7,
  xtick={-16,-8, ...,16},
  ymin=0, ymax=0.1,
  ytick={0, 0.02, 0.04, 0.06, 0.08, 0.1},
  minor y tick num=2,
  scaled y ticks=false,
  yticklabel style={
    /pgf/number format/precision=2,
    /pgf/number format/fixed,
    /pgf/number format/fixed zerofill,
  },
  tick label style={font=\tiny},
  xmajorgrids, ymajorgrids,
  axis background/.style={fill=white},
]
\addplot [
  red,
  only marks,
  mark=asterisk,
] table {figures/neighbour_distribution.dat};

\newcommand\B{5.6601}
\addplot [
  domain=-16:16,
  samples=33,
  blue,
  line width=0.8pt,
] {exp(-abs(x)/\B) / (2*\B)};
\end{axis}

\end{tikzpicture}
\caption{Distribution of the difference between neighboring pixel values of the standard test image ``Lenna" of size $512\times 512$.}
\label{fig:neighbourdist}
\end{figure}

Property~\ref{prop:neighbor} is well-known for natural images. Figure~\ref{fig:neighbourdist} shows the distribution of differences between neighboring pixel values of the test image ``Lenna" of size $512 \times 512$. With the real distribution (red), a Laplacian distribution $\mathrm{Laplace}(\mu,b)$ estimated from the real data is also shown, where the parameter $\mu$ is set to 0 and the bandwidth parameter $b$ was obtained by the maximum likelihood estimation algorithm.

Property~\ref{prop:range} is derived from the definition of the 2-D DCT. In the rest of the paper, we use $\bm{x}=\bm{A}\bm{y}$ to denote the relationship between image pixel values $\{x(i,j)\in[x_{\min}, x_{\max}]\}_{0\leq i,j \leq N-1}$ and DCT coefficients $\{y(k,l)\}_{0\leq k,l \leq N-1}$ in block-wise $N\times N$ 2-D DCT, which can be written as
\begin{equation}\label{eq:DCT2}
\begin{aligned}
 x(i,j) &= \sum_{0\leq k,l \leq N-1} A(i,j,k,l) \cdot y(k,l) \\
 &= \frac{1}{N}y(0,0) +\sum_{\substack{0 \leq k,l \leq N-1 \\ (k,l) \neq (0,0)}}A(i,j,k,l) \cdot y(k,l),
\end{aligned}
\end{equation}
where
\begin{equation*}
A(i,j,k,l)=C(k)C(l) \cos\left(\frac{(i+0.5)k\pi}{N}\right)
\cos\left(\frac{(j+0.5)l\pi}{N}\right),
\end{equation*}
and $C(k)$ is $\sqrt{1/N}$ when $k=0$ and $\sqrt{2/N}$ when $k>0$. As shown in Eq.~\eqref{eq:DCT2}, the DC coefficient $y(0,0)$ is constrained by the sum term without DC value and interval $[x_{\min}, x_{\max}]$. Note that Eq.~\eqref{eq:DCT2} is a linear equation and the indices are relative to each block.

In~\cite{Li:DCT:ICIP2011}, a similar DCT coefficient recovery problem is addressed, where missing DCT coefficients are estimated based on known ones. The main difference between the DCT coefficient recovery problem and the sign bit recovery problem is that the unknown variables in the latter are binary (1 or -1), but they are continuous in the former. Following the way how the DCT coefficient recovery problem is modeled, we can model the sign bit recovery problem as follows:
\begin{equation}
\label{eq:general}
\begin{aligned}
 \text{minimize }   & f(x)\\
 \text{subject to } & \bm{x}=\bm{A}\bm{y},\\
 & x_{\min} \leq x(i,j) \leq x_{\max},\\
 & y(k,l)=\begin{cases}
 y^*(k,l)       & \text{if the sign bit is known};\\
 s(y,l)|y(k,l)| & \text{if the sign bit is unknown}.
 \end{cases}\\
 & s(y,l)\in\{1, -1\},
\end{aligned}
\end{equation}
where $y^*(k,l)$ is the known DCT coefficient, $s(y,l)$ is the sign bit of the unknown DCT coefficient $y(k,l)$, and $f(x)$ is the objective function defined based on some properties of the image. We use the same objective function defined for the DCT coefficient recovery problem defined by Li \emph{et al.} in~\cite{Li:DCT:ICIP2011}:
\begin{equation}\label{eq:obj}
\begin{aligned}
\text{minimize }   & \sum{h_{i,j,i',j'}} \\
\text{subject to } & x(i,j) - x(i',j') \leq h_{i,j,i',j'}, \\
                   & x(i',j') - x(i,j) \leq h_{i,j,i',j'},
\end{aligned}
\end{equation}
where $(i,j)$ and $(i',j')$ are coordinates of neighboring pixels within the domain of the whole image. The above objective function is based on Property~\ref{prop:neighbor} and Fact~\ref{fact:Laplace}, and its actual effect is to define
the object function as
\[
f(x)=\sum_{i,j,i',j'}{|x(i,j)-x(i',j')|}.
\]
Equation~\eqref{eq:obj} uses a set of auxiliary variables $h_{i,j,i',j'}$ to linearize the above nonlinear objective function.

\begin{Fact}
\label{fact:Laplace}
Given $N$ observations $\{Z_i\}_{i=1}^N$ of a Laplacian distribution $\mathrm{Laplace}(\mu,b)$ with zero means $\mu=0$, the maximum likelihood estimator (MLE) of the parameter $b$ of the Laplacian distribution is $\frac{1}{N}\sum_{i=1}^N |Z_i|$.
\end{Fact}

The fact that the unknown variables $s(y,l)$ are binary means that the above model becomes a mixed integer linear programming (MILP) problem,\footnote{More specifically, our problem is a mixed binary integer linear programming, but its complexity is generally the same as the MILP, so in this paper, we just refer to it as a MILP problem.} rather than an LP problem for the DCT coefficient recovery problem.

\subsection{Two approximation methods}
\label{ssec:methods}

It has been known that as a general problem, the MILP problem is NP-hard so cannot be solved using a polynomial time algorithm. This means that we have to seek more efficient approximation methods. We propose two such methods, described below.

\subsubsection{Method 1--LP with relaxation}

We apply a linear relaxation to the constraint $y(k,l)=s(y,l)|y(k,l)|$ so that the DCT coefficient $y(k,l)$ with unknown sign bit can be assigned any value between $-|y(k,l)|$ and $|y(k,l)|$. In other words, the constraint becomes a normal LP problem's linear condition with upper and lower bounds:
\[
-|y(k,l)| \leq y(k,l) \leq |y(k,l)|.
\]
This change converts the MILP problem to a standard LP problem that can be solved in polynomial time. After the estimated DCT coefficient $\hat{y}(k,l)$ is obtained, we take its sign bit to determine the value of $s(y,l)$ and the final estimation of the coefficient as $\tilde{y}(k,l)=\text{sign}(\hat{y}(k,l))\cdot|y(k,l)|$, where the function $\text{sign}(\cdot)$ extracts the sign of a real number as a value 1 or $-1$. Note that if $\hat{y}(k,l)=0$ its sign bit is undefined, so we need a strategy to handle such a case. Four possible strategies are: 1) setting the coefficient to zero; 2) always assigning 1; 3) always assigning -1; 4) randomly assigning two possible sign bit values by following the Bernoulli distribution (i.e., assigning 1 with probability $p$ and $-1$ with probability $q=1-p$).

\subsubsection{Method 2--Hierarchical MILP or Hybrid MILP and LP}

In this method, we adopt a ``divide-and-conquer" (DAC) strategy to reduce the time complexity of the overall MILP problem without significantly compromising the visual quality of recovered images. The main idea is to reduce the MILP problem of the whole image into some MILP problems of smaller regions, and then to solve a smaller global MILP problem or a global LP problem to align the results of all the smaller region-wise MILP problems by refining DC coefficients (brightness) of all blocks in all regions. The regions need to be sufficiently small to make the smaller region-wise MILP problems solvable with practically small-time complexity. This DAC strategy could work because other than pixels on the boundary of each region, the smaller MILP problem should still be able to accurately recover all inner pixel values. The less accurate pixel values on the boundary can then be partially fixed using the final global optimization step.

The global LP step is done by fixing estimated AC coefficients and focusing only on adjusting the values of all blocks' DC coefficients. In other words, this step aims to globally align the brightness of all blocks in all regions such that the resulting image is as smooth as possible. We have two strategies for this step: 1) allowing different blocks within each region to have different DC coefficients; 2) assigning the same DC coefficient to all blocks in the same region, since the internal smoothness of each region should have been addressed by solving the corresponding region-specific MILP problem. These two strategies are called ``block LP" and ``region LP", respectively.

Normally, MILP problems are solved using a branch-and-bound algorithm, whose worst-case complexity is simply the size of the solution space. Assuming we have $u$ unknown sign bits, the worst-case complexity will be $O(2^u)$. Now, assuming we divide an $H \times W$ image into regions of fixed size $H'\times W'$ and the $u$ unknown sign bits are distributed uniformly across all regions, the $m=\frac{HW}{H'W'}$ smaller MILP problems will have an overall worst-case complexity of $O\left(2^{\frac{u}{m}}m\right)$. The global MILP problem will have a complexity of $O(2^{HW/N^2})$, and the global LP problem will have a polynomial-time complexity of $O\left(\left(HW/N^2\right)^{1.5}\right)$ (block LP) or $O\left(m^{1.5}\right)$ (region LP) if the most efficient known LP solving algorithm is used. By fixing $H'\times W'$ to be a sufficiently small size, e.g., $64 \times 64$ or $32\times 256$, we can effectively control the overall complexity of the whole process to be effectively polynomial time.

\subsection{Extension to different image encoding methods and settings}
\label{ssec:ext}

In real-world applications, digital images are always encoded following a specific image encoding standard such as JPEG, PNG, and GIF. Adopting JPEG, one of the most widely used image encoding standards, as an example, we also considered the following four special encoding settings of JPEG images in our approximation methods for solving the optimization problem:
\begin{itemize}
\item \textit{DC level shifting:} From Eq.~\eqref{eq:DCT2}, one can see that DC coefficients are always non-negative since they represent the average brightness of a block, which will be always non-negative. To more effectively encode DC coefficients, in JPEG, pixel values are first subtracted by half of the range (e.g., 128 for 8-bit images) before the 2-D DCT is applied. In this case, the sign bit of a DC coefficient can be positive or negative.

\item \textit{Quantization of DCT coefficients:}
Each coefficient $y(k,l)$ is divided by a quantized step and rounded to the nearest integer. Due to the error caused by quantization, we may need to relax the variable bounds further to make reasonable predictions. The relaxation for coefficient $y(k,l)$ is
\begin{equation}\label{eq:relax}
-Q(k,l)/2 \leq Y(k,l)\cdot Q(k,l) - y(k,l) \leq Q(k,l)/2,
\end{equation}
where $Y(k,l)$ is the quantized DCT coefficient and $Q(k,l)$ denotes the corresponding quantization step defined in the quantization table. The quantization of DCT coefficients means the range of pixel values calculated from such coefficients may go outside the valid range (e.g., [0,255] for 8-bit images), therefore, we need to consider how to relax the lower and upper bounds of pixel values by considering the effect of quantization errors. Considering the extreme case that quantization errors of all DCT coefficients have the same sign bit as the corresponding element in the 2-D DCT matrix $\bm{A}$, the maximum quantization error of the pixel at entry $(i,j)$ can be defined as
\begin{equation*}
\epsilon(i,j)=\sum_{0\leq k, l\leq N-1} \lvert A(i,j,k,l) \rvert \cdot Q(k,l)/2.
\end{equation*}
Considering the additional quantization errors that can be introduced in the solving process of the LP and MILP algorithm, we increase $\epsilon(i,j)$ by 1 to ensure such additional errors will still be tolerated. As a whole, the consideration of such quantization errors leads to a new condition for $x(i,j)$:
\[
x_{\min} - (\epsilon(i,j)+1) \leq x(i,j) \leq x_{\max} + (\epsilon(i,j)+1).
\]

\item \textit{Two's complement encoding of the DCT coefficients:}
Another coding feature in JPEG is the encoding of DCT coefficients using a standard table similar to the two's complement format. Accordingly, in the LP with the relaxation method, the relaxed upper and lower bounds of a coefficient with an unknown sign bit will be asymmetric since the non-sign bits are encoded differently for positive and negative values.

\item \textit{Encoding of DC coefficients:}
Each DC coefficient is encoded following the differential pulse code modulation method, i.e., what is encoded is the difference between the current DC coefficient and a previously encoded one. There are several ways to define the previous DC coefficient:\footnote{JPEG uses the first two, but some other image and video encoding standards use the third one as well, so in our paper, we consider all three.}
\begin{itemize}
\item \textit{DC prediction mode 1}: the previously coded block in the same row;

\item \textit{DC prediction mode 2}: the previously coded block as scanned in the raster order;

\item \textit{DC prediction mode 3}: the average of two previously coded blocks immediately above and/or to the left, scanned in the raster order.
\end{itemize}    
\end{itemize}

To deal with the above encoding details on the DC differential, a new variable $z$ is introduced to represent the result. Since we only know the absolute value of an encoded difference $z$, the constraint on $z$ is $z\in\{-|z|, |z|\}$, and the corresponding linear relaxation is $-|z|\leq z\leq |z|$. The relationship between $z$ and the DC coefficient $y(0,0)$ is defined according to the differential encoding scheme. For example, if prediction mode 1 is adopted, one can deduce $y(0,0)-y'(0,0)=z$, where $y'(0,0)$ is the DC coefficient of the previous block in the same row. The DC coefficient $y(0,0)$ is determined by the variable $z$. For comparison, we also define \textit{DC prediction mode 0} to be the case where the original DC coefficients are encoded (i.e., without using the differential encoding scheme). Moreover, the dependency introduced by the differential encoding may cause error propagation to occur. This can seriously downgrade the visual quality of restored images.

When the region-wise MILP method is used with DC differential encoding, some DC coefficients of the current region are dependent on DC coefficients from one or more previous regions. We propose the following three different strategies to solve such inter-regional dependencies.
\begin{itemize}
\item \textit{Dependency mode 0} -- Removing the dependency completely. Then, the corresponding encoded difference is not exploited, although the dependency within the region is still maintained.

\item \textit{Dependency mode 1} -- Solving all regions following the raster order so that all previously encoded DC coefficients are available for the current region. For DC prediction mode 2, the leftmost block of each row relies on the rightmost block of the above row, so if the region covers more than one block row, the region has to be as wide as the whole image. Otherwise, some regions to the left will have dependencies on some other regions to the right that have not been previously solved. In other words, the valid region size of this and the next modes can only be $N \times k_1N$ or $k_2N \times W$, where $k_1 \geq 1$ and $k_2>1$. In this mode, the MILP problem of each region is limited to pixel values and DCT coefficients of the current region.

\item \textit{Dependency mode 2} -- The same as dependency mode 1 except for each MILP problem of a region we also consider pixel value differences between the current region and the one or more adjacent region(s) that have been previously solved. Since DCT coefficients and pixel values of all previously solved regions are already solved, they can be considered known so that the size of the MILP problem remains the same in terms of the number of unknown variables.
\end{itemize}

\section{Experimental Results}
\label{sec:results}



In this section, we first give the results of the extensive experiments conducted on 30 typical images (22 of size $256 \times 256$ and 8 of size $384 \times 256$) to verify the performance of the two recovery methods. The objective metrics PSNR and SSIM were used as the main quantitative indicators for evaluating the visual quality of reconstructed images. We also conducted subjective quality evaluations of selected images to ensure the objective quality indicators match the actual visual quality perceived by us as expert observers. Then, we briefly compare our methods with some naive recovery methods. Note that all experiments were implemented in Matlab with IBM\textsuperscript{\textregistered} ILOG\textsuperscript{\textregistered} CPLEX\textsuperscript{\textregistered}.\footnote{\url{https://www.ibm.com/products/ilog-cplex-optimization-studio}}

\subsection{Performance of the relaxed LP method}

In this subsection, we show our thorough analysis of the impact of several key parameters and different implementation strategies on the recovery performance of the relaxed LP method.

\subsubsection{Accelerating computation by ignoring small DCT coefficients}

To speed up the computation by reducing the size of the solution space, we directly set an unknown DCT coefficient to zero if its absolute value is lower than a certain threshold $T$. Since the coefficients close to zero contribute less to pixel values, ignoring them may not affect the visual quality of recovered results much. To find an appropriate threshold that balances the computational complexity and the recovery quality, we evaluated the computational time and visual quality of results at various threshold values under different DC prediction modes.

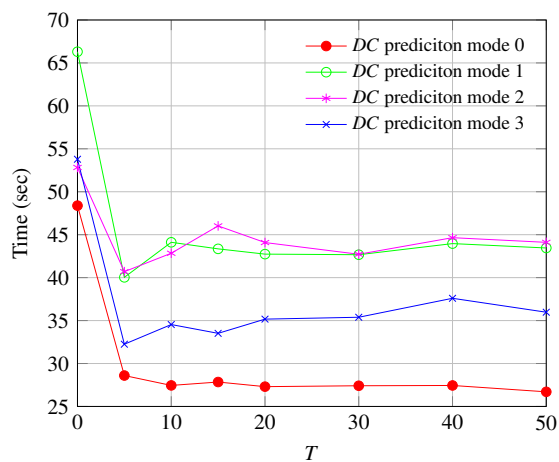
\begin{figure}[!htb]
\centering
\begin{tikzpicture}[scale=0.9]
\begin{axis}[
xlabel={$T$},
ylabel={Time (sec)},
xmin=0, xmax=50,
xtick={0, 10, ..., 50},
ymin=25, ymax=70,
ytick={25, 30, ..., 75},
xmajorgrids, ymajorgrids,
]
\addplot [red, mark=*] table[row sep=\\]{
0	48.383\\
5	28.6\\
10	27.449\\
15	27.844\\
20	27.306\\
30	27.41\\
40	27.443\\
50	26.703\\
};
\addlegendentry{$DC$ prediciton mode 0}

\addplot [green, mark=o] table[row sep=\\]{
0	66.303\\
5	40.033\\
10	44.114\\
15	43.349\\
20	42.737\\
30	42.668\\
40	43.963\\
50	43.453\\
};
\addlegendentry{$DC$ prediciton mode 1}

\definecolor{temp}{rgb}{1,0,1}
\addplot [temp, mark=asterisk] table[row sep=\\]{
0	52.825\\
5	40.695\\
10	42.839\\
15	46.022\\
20	44.085\\
30	42.721\\
40	44.652\\
50	44.108\\
};
\addlegendentry{$DC$ prediciton mode 2}

\addplot [blue, mark=x] table[row sep=\\]{
0	53.785\\
5	32.248\\
10	34.537\\
15	33.513\\
20	35.167\\
30	35.387\\
40	37.594\\
50	35.973\\
};
\addlegendentry{$DC$ prediciton mode 3}
\end{axis}
\end{tikzpicture}
\caption{Time consumption under different values of threshold.}
\label{fig:Ttime}
\end{figure}

\begin{figure*}[!htb]
\centering
\subfloat[]{\begin{tikzpicture}[scale=0.9]
\begin{axis}[
  xlabel={$T$},
  ylabel={PSNR},
  xmin=0, xmax=50,
  ymin=11, ymax=23,
  xtick={0, 10, ..., 50},
  ytick={11, 12.5, ..., 23},
  xmajorgrids, ymajorgrids,
  legend style={at={(0.37,0.61)}, anchor=south west}
]
\addplot [red, mark=*] table [row sep=\\] {
0   21.6392\\
5   21.8595\\
10  21.5981\\
15  21.717\\
20  21.8617\\
30  22.0791\\
40  22.032\\
50  21.8326\\
};
\addlegendentry{$DC$ prediciton mode 0}

\addplot [green, mark=o] table[row sep=\\]{
0 15.7165\\
5 15.5691\\
10  15.5744\\
15  15.6319\\
20  15.6104\\
30  16.2311\\
40  16.108\\
50  15.8158\\
};
\addlegendentry{$DC$ prediciton mode 1}

color=\definecolor{temp}{rgb}{1,0,1}
\addplot [temp, mark=asterisk] table[row sep=\\]{
0 11.9619\\
5 12.1725\\
10  11.9081\\
15  12.4984\\
20  12.5254\\
30  12.4837\\
40  12.636\\
50  12.6914\\
};
\addlegendentry{$DC$ prediciton mode 2}

\addplot [color=blue, mark=x] table[row sep=\\]{
0 17.4759\\
5 17.7858\\
10  17.8022\\
15  17.805\\
20  17.8144\\
30  18.062\\
40  18.4374\\
50  18.0094\\
};
\addlegendentry{$DC$ prediciton mode 3}
\end{axis}
\end{tikzpicture}}
\quad
\subfloat[]{\begin{tikzpicture}[scale=0.9]
\begin{axis}[
xlabel={$T$},
ylabel={SSIM},
xmin=0, xmax=50,
xtick={0, 10, ..., 50},
ymin=0.5, ymax=0.9,
ytick={0.5, 0.55, ..., 0.91},
xmajorgrids, ymajorgrids,
legend style={at={(0.37,0.42)}, anchor=south west}
]
\addplot [red, mark=*] table[row sep=\\]{
0	0.832703\\
5	0.839644\\
10	0.837797\\
15	0.839865\\
20	0.840924\\
30	0.839675\\
40	0.837533\\
50	0.830557\\
};
\addlegendentry{$DC$ prediciton mode 0}

\addplot [green, mark=o] table[row sep=\\]{
0	0.652918\\
5	0.656716\\
10	0.655018\\
15	0.660617\\
20	0.666051\\
30	0.66245\\
40	0.660082\\
50	0.65231\\
};
\addlegendentry{$DC$ prediciton mode 1}

\definecolor{temp}{rgb}{1,0,1}
\addplot [temp, mark=asterisk] table[row sep=\\]{
0	0.557779\\
5	0.564832\\
10	0.556123\\
15	0.568051\\
20	0.576661\\
30	0.559431\\
40	0.563826\\
50	0.552831\\
};
\addlegendentry{$DC$ prediciton mode 2}

\addplot [blue, mark=x] table[row sep=\\]{
0	0.782082\\
5	0.781984\\
10	0.784886\\
15	0.789442\\
20	0.792495\\
30	0.798762\\
40	0.797086\\
50	0.784168\\
};
\addlegendentry{$DC$ prediciton mode 3}
\end{axis}
\end{tikzpicture}}
\caption{Quality of the recovery result under different values of threshold $T$: a) PSNR; b) SSIM.}
\label{fig:TPerformance}
\end{figure*}
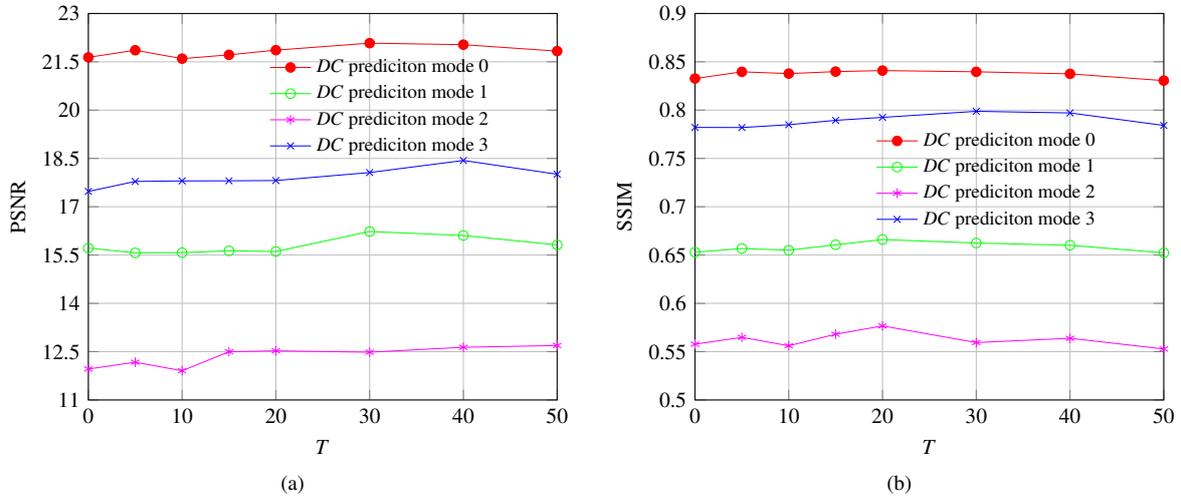

When there are fewer missing coefficients, we found that the speed acceleration effect is weak because only a few coefficients have closer-to-zero values. Increasing the number of missing coefficients $U$ in the ``zigzag" order used in the JPEG standard, we found that the speedup became more apparent and observed a sharp reduction in the computation time at a specific threshold. As shown in Fig.~\ref{fig:Ttime}, the time reduction happens at $T=5$ when $U=6$. The corresponding recovery quality at various threshold values is shown in Fig.~\ref{fig:TPerformance}. It can be seen that the performance is not very sensitive to the change of threshold $T$. According to the experimental results, we set $T=5$ for other experiments, which can provide speedup to some extent while preserving almost the same recovery performance.

\subsubsection{How the DC encoding scheme affects the recovery result}

\begin{figure}[!htb]
\centering
\subfloat[]{\includegraphics[width=\twofigwidth]{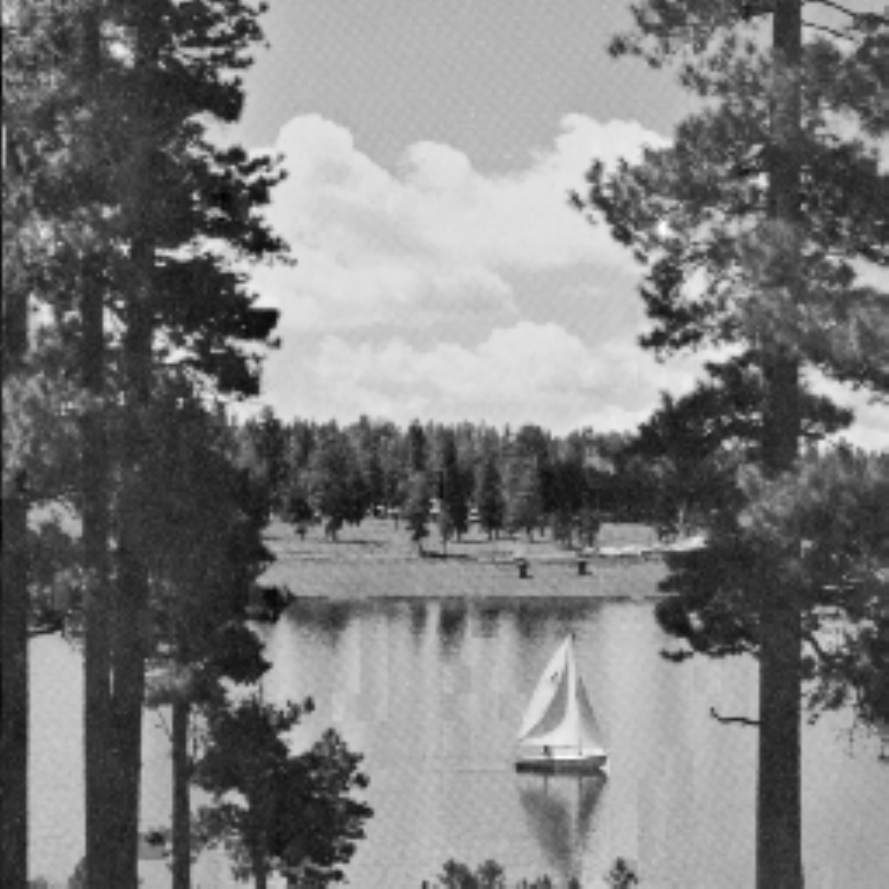}}
\quad
\subfloat[]{\includegraphics[width=\twofigwidth]{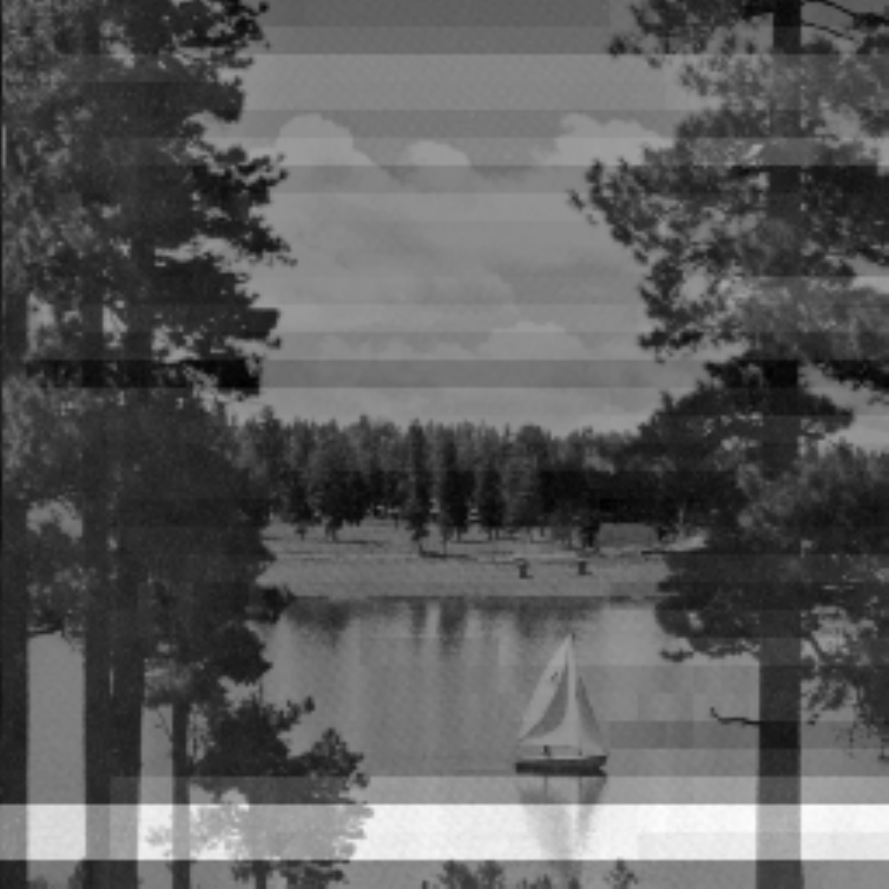}}
\\
\subfloat[]{\includegraphics[width=\twofigwidth]{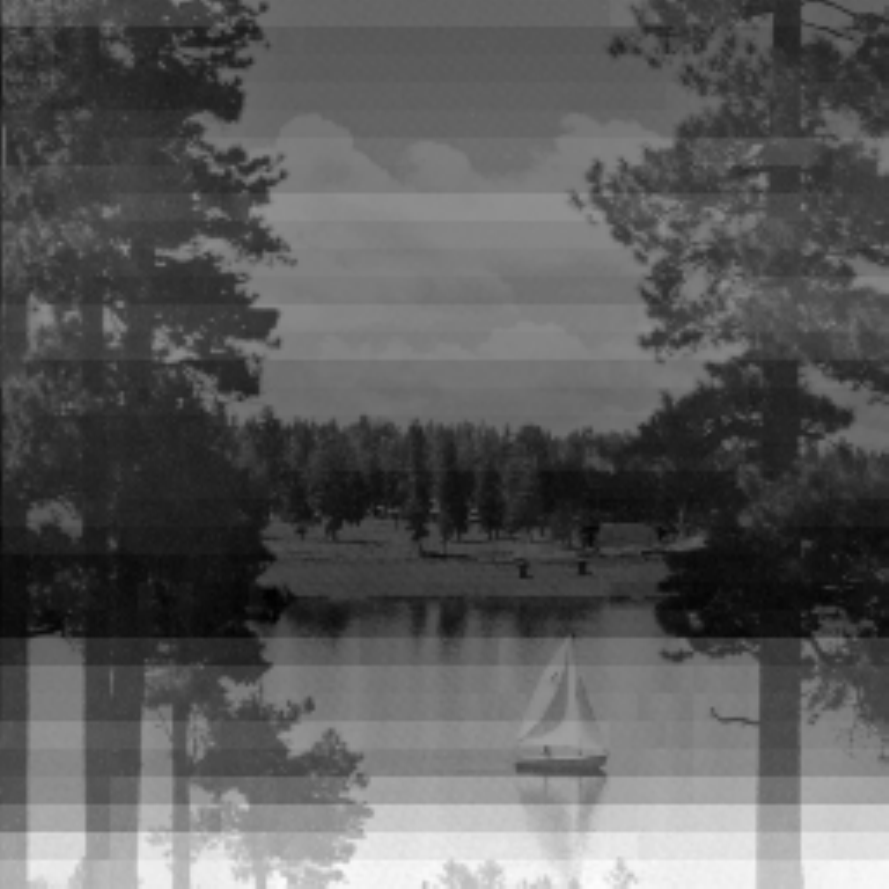}}
\quad
\subfloat[]{\includegraphics[width=\twofigwidth]{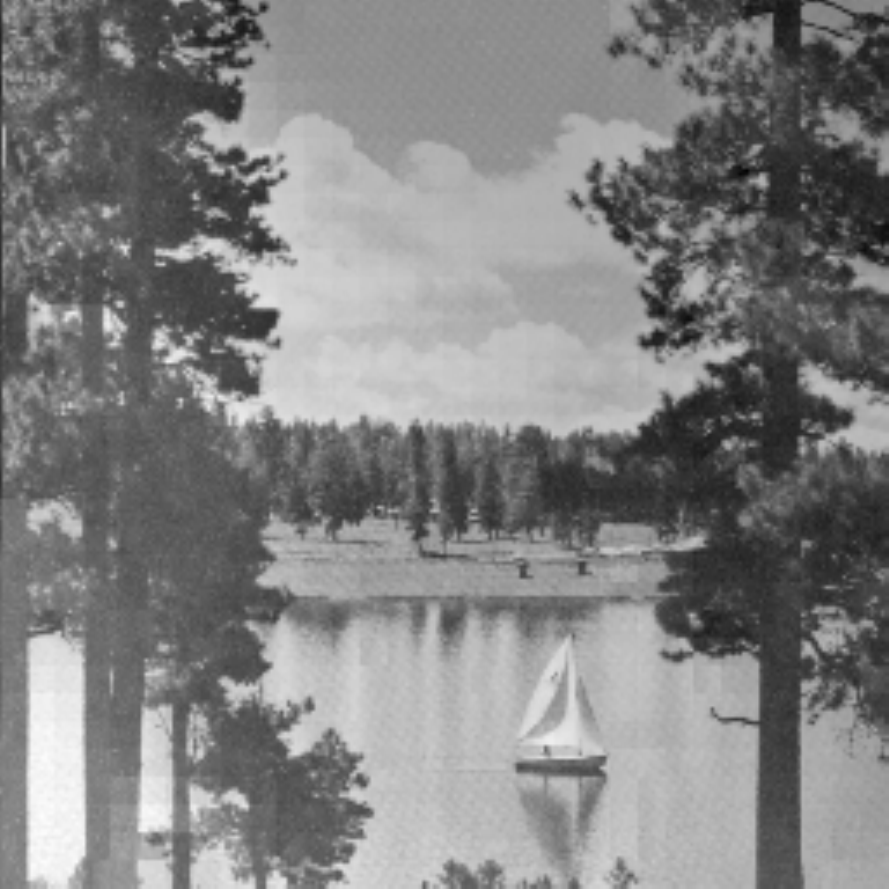}}
\caption{Recovered images under different DC prediction modes ($U=2$):
a) mode 0; b) mode 1; c) mode 2; d) mode 3.}
\label{fig:DCpred}
\end{figure}

The DC differential encoding can impact the recovery performance because of the error propagation effect. Table~\ref{table:dcpredmode} lists the recovery performance of the four different DC prediction modes with $U=2$. Observing the results in Fig.~\ref{fig:TPerformance}, one can see that the best performance was obtained in DC prediction mode 0, i.e., when the differential encoding is absent. This is consistent with our expectation since there is no error propagation effect in mode 0. In other modes, if an error occurs for one block, then the following blocks depending on it for DC prediction will be affected. As the error propagates over the entire image in mode 2, the recovery performance is worst among the four cases. Mode 1 is slightly better because the error propagates only within each row. Although the error also propagates globally in mode 3, the visual quality of recovered images is superior to that of modes 1 and 2. This may be attributed to the fact that the DC prediction is dependent on the block just above and to the left of the current block. So the error propagation effect might be alleviated by taking the average of the DC coefficients of the two blocks. To visualize the impact of the aforementioned error propagation effect, we display four recovered images under the four prediction modes in Fig.~\ref{fig:DCpred}.

\begin{table}[!htb]
\centering
\caption{Performance comparison under different DC prediction modes ($U=2$).}
\label{table:dcpredmode}
\begin{tabular}{*{4}{c|}c}
\hline
\multirow{2}{*}{Prediction mode} & \multicolumn{2}{c|}{SSIM} & \multicolumn{2}{c}{PSNR}\\
\cline{2-5}
& mean & median & mean & median\\
\hline
0 & 0.966288 & 0.973865 & 29.7510 & 29.3294\\
1 & 0.837763 & 0.853510 & 19.8658 & 20.0097\\
2 & 0.747883 & 0.745756 & 14.9363 & 14.0246\\
3 & 0.933148 & 0.950445 & 23.2667 & 23.0676\\
\hline
\end{tabular}
\end{table}

\subsubsection{How the sign bit recovery method affects the recovery result}

After solving the linear relaxation of the original problem, one needs to convert the obtained coefficients to sign bits. A simple way is to directly extract the sign bits of estimated coefficients. However, this leads to ambiguity when the obtained coefficients happen to be zero. In such cases, we adopt the following four methods: 1) setting the coefficient to zero by ignoring the known absolute value of the DCT coefficient; 2) setting the sign to 1; 3) setting the sign to -1; and 4) assigning the sign randomly.

The recovery performances of the above four methods are listed in Table~\ref{table:recovery_methods} when $U=2$ and with the DC differential encoding disabled (i.e., DC prediction mode 0). As shown in Table~\ref{table:recovery_methods}, the performances of the first four methods are very similar, suggesting that different mapping methods have little impact on the visual quality of the final recovered images. The results are similar for different values of $U$ and prediction modes of DC, too. Based on the results, we adopted method 1 for other experiments.

\begin{table}[!htb]
\centering
\caption{Performance comparison using different sign bit recovery methods.}
\label{table:recovery_methods}
\begin{tabular}{*{4}{c|}c}
\hline
\multirow{2}{*}{Recovery method} & \multicolumn{2}{c|}{SSIM} & \multicolumn{2}{c}{PSNR} \\ \cline{2-5}
& mean & median & mean & median \\
\hline
1 & 0.966288 & 0.973865 & 29.7510 & 29.3294 \\
2 & 0.966101 & 0.973961 & 29.7666 & 29.3284 \\
3 & 0.966097 & 0.973337 & 29.7318 & 29.3284 \\
4 & 0.966134 & 0.973667 & 29.7492 & 29.3287 \\
\hline
\end{tabular}
\end{table}

\subsubsection{How the bound relaxation affects the recovery result}

As mentioned before, due to the way that DCT coefficients of JPEG images are encoded, we may need to further relax the ranges of pixels and coefficients to cope with the quantization error. Let $R_x$ and $R_y$ denote whether relax $x$ and $y$, respectively. Table~\ref{table:relax} exhibits the recovery performance of four different combinations of bound relaxations, when $U=2$ and the JPEG quality factor (QF) is set to 95. Here, we only consider the DC prediction modes 1 and 2 used in the JPEG coding. As shown in Table~\ref{table:relax}, adding extra relaxations does not improve the performance, and even degrades the performance in some cases. Therefore, the additional relaxation looks unnecessary.

\setlength\tabcolsep{3pt} 
\newcommand{\minitab}[2][l]{\begin{tabular}{#1}#2\end{tabular}}
\begin{table}[!htb]
\centering
\caption{Performance comparison for bound relaxations.}
\label{table:relax}
\begin{tabular}{*{5}{c|}c}
\hline
\multirow{2}*{\minitab[c]{Prediction\\ mode}}
& \multirow{2}{*}{Parameter} & \multicolumn{2}{c|}{SSIM} & \multicolumn{2}{c}{PSNR}\\
\cline{3-6}
& & mean & median & mean & median\\
\hline
\multirow{4}{*}{1}
& $R_x=0$, $R_y=0$ & 0.816792 & 0.832737 & 19.4089 & 19.0661\\
& $R_x=0$, $R_y=1$ & 0.808481 & 0.830511 & 19.0194 & 18.7751\\
& $R_x=1$, $R_y=0$ & 0.809730 & 0.832970 & 19.2732 & 18.9590\\
& $R_x=1$, $R_y=1$ & 0.803481 & 0.831153 & 18.8007 & 19.0441\\
\hline
\multirow{4}{*}{2}
& $R_x=0$, $R_y=0$ & 0.714905 & 0.724600 & 14.2486 & 14.2310\\
& $R_x=0$, $R_y=1$ & 0.703095 & 0.714658 & 14.0441 & 13.2331\\
& $R_x=1$, $R_y=0$ & 0.699856 & 0.709658 & 13.7694 & 13.7121\\
& $R_x=1$, $R_y=1$ & 0.692502 & 0.701389 & 13.6705 & 12.9393\\
\hline
\end{tabular}
\end{table}

Without the above relaxation, the coefficient $y$ and its quantized version $Y$ are linearly correlated since $y(k,l)=Y(k,l)\cdot qt(k,l)$. However, this relationship does not exist when $R_y=1$ according to Eq.~\eqref{eq:relax}. Then the number of variables to be determined and time consumption increase considerably. Table~\ref{table:relax_time} shows the time consumption in the above-relaxed optimization. One can see that relaxing $x$ does not affect the computation efficiency much, but relaxing $y$ increases the computation time by nearly five times. The comparison results of time consumption and performance for different $U$ are quite similar. Based on the results, we set $R_x=1$ and $R_y=0$ for the subsequent experiments on JPEG images.

\begin{table}[!htb]
\centering
\caption{Speed comparison for bound relaxations.}
\label{table:relax_time}
\begin{tabular}{c|c|c}
\hline
Prediction mode & Parameter & Time (mean) (sec)\\
\hline
\multirow{4}{*}{1}
& $R_x=0$, $R_y=0$ & 42.094\\
& $R_x=1$, $R_y=0$ & 40.338\\
& $R_x=0$, $R_y=1$ & 204.371\\
& $R_x=1$, $R_y=1$ & 181.869\\
\hline
\multirow{4}{*}{2}
& $R_x=0$, $R_y=0$ & 42.282\\
& $R_x=1$, $R_y=0$ & 41.649\\
& $R_x=0$, $R_y=1$ & 176.555\\
& $R_x=1$, $R_y=1$ & 208.734\\
\hline
\end{tabular}
\end{table}


\subsubsection{How the JPEG quality factor affects the recovery result}


When the JPEG quality factor decreases, DCT coefficients are divided by larger quantized values in quantization, which causes more known coefficients to be zero. Zero-value coefficients provide less useful information for optimization, so the recovery performance becomes worse. Tables~\ref{table:qf_1} and \ref{table:qf_2} display the recovery performance on JPEG images in DC prediction modes 1 and 2, respectively. In most cases, the recovery performance only drops slightly as the JPEG quality factor decreases. As the performance drop is relatively small, the optimization method can still handle JPEG images well.

\begin{table}[!htb]
\center
\caption{Performance comparison under different quality factors in DC prediction mode 1.}
\label{table:qf_1}
\begin{tabular}{*{5}{c|}c}
\hline
\multirow{2}{*}{$U$} & \multirow{2}{*}{QF} & \multicolumn{2}{c|}{SSIM} & \multicolumn{2}{c}{PSNR}\\
\cline{3-6}
& & mean & median & mean & median\\
\cline{1-6}
\multirow{3}{*}{1}
& 75 & 0.867648 & 0.909504 & 21.6592 & 21.7493\\
& 85 & 0.883085 & 0.910543 & 23.1157 & 23.9398\\
& 95 & 0.889687 & 0.914645 & 22.2965 & 22.8775\\
\hline
\multirow{3}{*}{2}
& 75 & 0.778408 & 0.785937 & 17.9644 & 17.6650\\
& 85 & 0.801444 & 0.825158 & 18.6412 & 18.5521\\
& 95 & 0.808481 & 0.830511 & 19.0194 & 18.7751\\
\hline
\multirow{3}{*}{4}
& 75 & 0.658763 & 0.666987 & 15.6341 & 15.4466\\
& 85 & 0.679338 & 0.662292 & 15.7255 & 15.6664\\
& 95 & 0.698067 & 0.693938 & 15.8881 & 15.9850\\
\hline
\multirow{3}{*}{8}
& 75 & 0.526632 & 0.508482 & 13.8857 & 13.9358\\
& 85 & 0.568085 & 0.559700 & 14.2235 & 13.7504\\
& 95 & 0.569394 & 0.563065 & 14.2680 & 14.3125\\
\hline
\end{tabular}
\end{table}

\begin{table}[!htb]
\center
\caption{Performance comparison under different quality factors in DC prediction mode 2.}
\label{table:qf_2}
\begin{tabular}{*{5}{c|}c}
\hline
\multirow{2}{*}{$U$} & \multirow{2}{*}{QF} & \multicolumn{2}{c|}{SSIM} & \multicolumn{2}{c}{PSNR}\\
\cline{3-6}
&  & mean & median & mean & median\\
\cline{1-6}
\multirow{3}{*}{1}
& 75 & 0.819763 & 0.847841 & 17.4260 & 17.0793\\
& 85 & 0.820909 & 0.842643 & 17.6470 & 17.2557\\
& 95 & 0.816946 & 0.838111 & 17.4661 & 16.9942\\
\hline
\multirow{3}{*}{2}
& 75 & 0.672748 & 0.677340 & 13.6551 & 13.1976\\
& 85 & 0.693293 & 0.710507 & 13.9767 & 13.0946\\
& 95 & 0.703095 & 0.714658 & 14.0441 & 13.2331\\
\hline
\multirow{3}{*}{4}
& 75 & 0.543130 & 0.554039 & 11.3950 & 10.8837\\
& 85 & 0.579811 & 0.546006 & 12.3227 & 11.3719\\
& 95 & 0.602612 & 0.583554 & 12.8899 & 12.5304\\
\hline
\multirow{3}{*}{8}
& 75 & 0.436596 & 0.436948 & 10.9182 & 10.9728\\
& 85 & 0.455535 & 0.439074 & 11.0495 & 10.9998\\
& 95 & 0.481800 & 0.457175 & 11.5209 & 11.0480\\
\hline
\end{tabular}
\end{table}


\subsubsection{How the number of missing sign bits affects the recovery result}

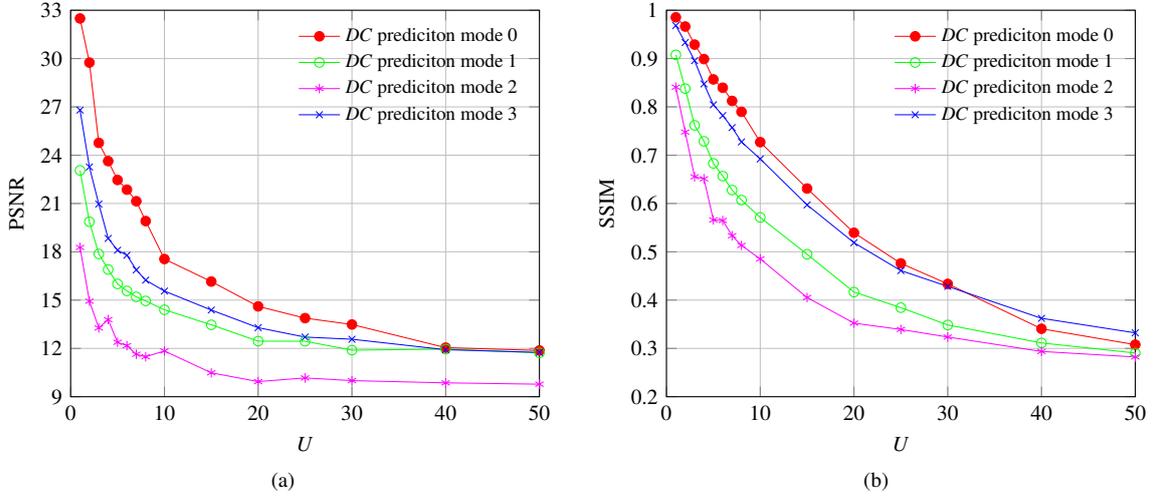
\begin{figure*}[!htb]
\centering
\subfloat[]{\begin{tikzpicture}[scale=0.9]
\begin{axis}[
xlabel={$U$},
ylabel={PSNR},
xmin=0, xmax=50,
xtick={0, 10, ..., 50},
ymin=9, ymax=33,
ytick={9, 12, ..., 33},
xmajorgrids, ymajorgrids,
]
\addplot [red, mark=*] table[row sep=\\]{
1	32.4904\\
2	29.751\\
3	24.7634\\
4	23.6314\\
5	22.4608\\
6	21.8595\\
7	21.1414\\
8	19.9078\\
10	17.5505\\
15	16.1521\\
20	14.6108\\
25	13.8813\\
30	13.4849\\
40	12.0481\\
50	11.8731\\
};
\addlegendentry{$DC$ prediciton mode 0}

\addplot [green, mark=o] table[row sep=\\]{
1	23.0593\\
2	19.8658\\
3	17.8673\\
4	16.8996\\
5	16.007\\
6	15.5691\\
7	15.2008\\
8	14.9471\\
10	14.3997\\
15	13.4648\\
20	12.4511\\
25	12.4458\\
30	11.8911\\
40	11.9573\\
50	11.726\\
};
\addlegendentry{$DC$ prediciton mode 1}

\definecolor{temp}{rgb}{1,0,1}
\addplot [temp, mark=asterisk] table[row sep=\\]{
1	18.2705\\
2	14.9363\\
3	13.271\\
4	13.7868\\
5	12.3805\\
6	12.1725\\
7	11.6365\\
8	11.4825\\
10	11.8423\\
15	10.4848\\
20	9.9387\\
25	10.1668\\
30	9.9999\\
40	9.8608\\
50	9.7774\\
};
\addlegendentry{$DC$ prediciton mode 2}

\addplot [blue, mark=x] table[row sep=\\]{
1	26.8006\\
2	23.2667\\
3	20.9656\\
4	18.8303\\
5	18.0978\\
6	17.7858\\
7	16.8749\\
8	16.24\\
10	15.5621\\
15	14.3832\\
20	13.2884\\
25	12.7039\\
30	12.5698\\
40	11.9202\\
50	11.7503\\
};
\addlegendentry{$DC$ prediciton mode 3}

\end{axis}
\end{tikzpicture}}
\quad
\subfloat[]{\begin{tikzpicture}[scale=0.9]
\begin{axis}[
xlabel={$U$},
ylabel={SSIM},
xmin=0, xmax=50,
xtick={0, 10, ..., 50},
ymin=0.2, ymax=1,
ytick={0.2, 0.3,..., 1},
xmajorgrids, ymajorgrids,
]
\addplot [red, mark=*] table[row sep=\\]{
1	0.985425\\
2	0.966288\\
3	0.928892\\
4	0.89893\\
5	0.8569\\
6	0.839644\\
7	0.812352\\
8	0.789797\\
10	0.727012\\
15	0.630975\\
20	0.539432\\
25	0.476044\\
30	0.433552\\
40	0.340651\\
50	0.307773\\
};
\addlegendentry{$DC$ prediciton mode 0}

\addplot [green, mark=o] table[row sep=\\]{
1	0.907737\\
2	0.837763\\
3	0.761742\\
4	0.728705\\
5	0.683058\\
6	0.656716\\
7	0.627775\\
8	0.607095\\
10	0.571174\\
15	0.49489\\
20	0.416642\\
25	0.384399\\
30	0.348542\\
40	0.311339\\
50	0.290469\\
};
\addlegendentry{$DC$ prediciton mode 1}

\definecolor{temp}{rgb}{1,0,1}
\addplot [temp, mark=asterisk] table[row sep=\\]{
1	0.840738\\
2	0.747883\\
3	0.655252\\
4	0.651085\\
5	0.565992\\
6	0.564832\\
7	0.533264\\
8	0.513127\\
10	0.48529\\
15	0.405333\\
20	0.352289\\
25	0.339229\\
30	0.323592\\
40	0.293723\\
50	0.282281\\
};
\addlegendentry{$DC$ prediciton mode 2}

\addplot[blue, mark=x] table[row sep=\\]{
1	0.968316\\
2	0.933148\\
3	0.89567\\
4	0.847502\\
5	0.80431\\
6	0.781984\\
7	0.757118\\
8	0.727495\\
10	0.692179\\
15	0.597053\\
20	0.518858\\
25	0.461052\\
30	0.428364\\
40	0.362315\\
50	0.332213\\
};
\addlegendentry{$DC$ prediciton mode 3}
\end{axis}
\end{tikzpicture}}
\caption{Performance comparison under the different number of missing coefficients: a) PSNR; b) SSIM.}
\label{fig:u_pgm}
\end{figure*}

When more sign bits are missing, the optimization becomes more intractable and the recovery performance should degrade accordingly. We conducted experiments based on the assumption that in each DCT block, sign bits of the $U$ most significant DCT coefficients are missing. To facilitate comparison, we increased the number of missing sign bits in the ``zigzag" order used in JPEG. Since higher-frequency DCT coefficients have less energy statistically, we predicted that the reduction of recovery performance would get smaller as $U$ increases. Figure~\ref{fig:u_pgm} shows the mean visual quality of restored images given some missing coefficients and a DC prediction mode. The PSNR values of recovered images drop rapidly when $U \leq 10$, and then decreases relatively slowly. In terms of SSIM, the general trend is that the performance keeps decreasing at a moderate speed, and the slope gets smaller with larger $U$. We also conducted experiments on JPEG images with a QF of 95 and the corresponding results are shown in Table~\ref{table:u_jpg}. Similarly, the visual quality drops rapidly when $U \leq 8$, and then the trend gradually flattens.

\begin{table}[!htb]
\centering
\caption{Performance comparison under the different number of missing coefficients.}
\label{table:u_jpg}
\begin{tabular}{*{5}{c|}c}
\hline
\multirow{2}{*}{Prediction mode} & \multirow{2}{*}{U} & \multicolumn{2}{c|}{SSIM} & \multicolumn{2}{c}{PSNR}\\
\cline{3-6}
& & mean & median & mean & median\\
\hline
\multirow{5}{*}{1}
& 1 & 0.889687 & 0.914645 & 22.2965 & 22.8775\\
& 2 & 0.808481 & 0.830511 & 19.0194 & 18.7751\\
& 4 & 0.698067 & 0.693938 & 15.8881 & 15.9850\\
& 8 & 0.569394 & 0.563065 & 14.2680 & 14.3125\\
& 16 & 0.437425 & 0.398661 & 12.5036 & 12.4514\\
\hline
\multirow{5}{*}{2}
& 1 & 0.816946 & 0.838111 & 17.4661 & 16.9942\\
& 2 & 0.703095 & 0.714658 & 14.0441 & 13.2331\\
& 4 & 0.602612 & 0.583554 & 12.8899 & 12.5304\\
& 8 & 0.481800 & 0.457175 & 11.5209 & 11.0480\\
& 16 & 0.372325 & 0.320282 & 10.1647 & 9.9707\\
\hline
\end{tabular}
\end{table}

\subsubsection{Time consumption}

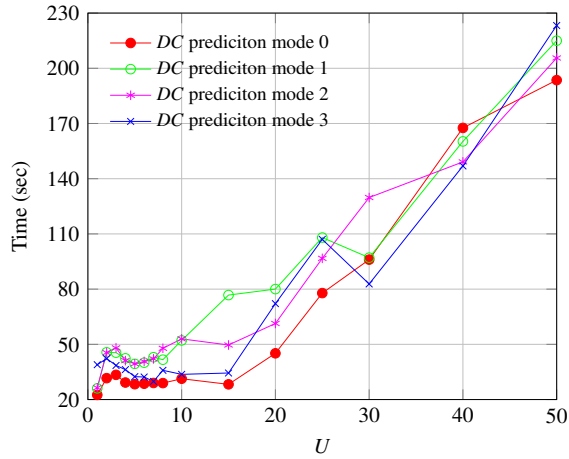
\begin{figure}[!htb]
\centering
\begin{tikzpicture}[scale=0.9]
\begin{axis}[
xlabel={$U$},
ylabel={Time (sec)},
xmin=0, xmax=50,
xtick={0, 10, ..., 50},
ymin=20, ymax=230,
ytick={20, 50, ..., 230},
xmajorgrids, ymajorgrids,
legend style={at={(0.03,0.97)}, anchor=north west}
]
\addplot [red, mark=*] table[row sep=\\]{
1	22.526\\
2	31.624\\
3	33.43\\
4	29.248\\
5	28.334\\
6	28.6\\
7	28.994\\
8	28.933\\
10	31.311\\
15	28.261\\
20	45.125\\
25	77.845\\
30	96.027\\
40	167.596\\
50	193.565\\
};
\addlegendentry{$DC$ prediciton mode 0}

\addplot [green, mark=o] table[row sep=\\]{
1	26.095\\
2	45.655\\
3	45.47\\
4	42.388\\
5	39.389\\
6	40.033\\
7	42.985\\
8	41.649\\
10	52.124\\
15	76.775\\
20	80.021\\
25	108.026\\
30	97.096\\
40	160.266\\
50	214.992\\
};
\addlegendentry{$DC$ prediciton mode 1}

\definecolor{temp}{rgb}{1,0,1}
\addplot [temp, mark=asterisk] table[row sep=\\]{
1	26.426\\
2	45.495\\
3	48.081\\
4	41.141\\
5	39.45\\
6	40.695\\
7	42.105\\
8	47.858\\
10	52.971\\
15	49.647\\
20	61.337\\
25	96.718\\
30	129.767\\
40	149.125\\
50	205.595\\
};
\addlegendentry{$DC$ prediciton mode 2}

\addplot [blue, mark=x] table[row sep=\\]{
1	38.937\\
2	42.263\\
3	38.522\\
4	36.271\\
5	32.438\\
6	32.248\\
7	29.847\\
8	35.876\\
10	33.691\\
15	34.423\\
20	72.122\\
25	107.036\\
30	82.806\\
40	146.931\\
50	223.251\\
};
\addlegendentry{$DC$ prediciton mode 3}

\end{axis}
\end{tikzpicture}
\caption{Time consumption with the different number of missing coefficients.}
\label{fig:u_t_pgm}
\end{figure}

Similar to~\cite{Li:DCT:ICIP2011}, the time complexity of the resulting linear relaxation problem is $O(n^4m^4U)$, where $n\times m$ is the size of the input image. Figure~\ref{fig:u_t_pgm} shows the average time consumption with respect to different numbers of missing coefficients, and the increase is roughly in line with the theoretical estimate. Besides, we found that the time consumption across different images can vary by several times. The optimization is particularly time-consuming for images with monotonous backgrounds. In these backgrounds, most of the coefficients are equal to zero and provide less useful information, which causes the optimization to be more intractable and therefore consume more time. As the quality factor descends, the computational time rises remarkably, which may be explained by the existence of more zero-value coefficients after quantization (less available information) to find the optimal solution. The time consumption on some JPEG images with various QF is reported in Table~\ref{table:time-jpg}.

\begin{table}[!htb]
\center
\caption{Time consumption on JPEG images.}
\label{table:time-jpg}
\begin{tabular}{*{3}{c|}c}
\hline
\multirow{2}{0.1in}{$U$} & \multirow{2}{0.2in}{QF} & \multicolumn{2}{c}{Time (sec)}\\ \cline{3-4}
& & mode 1 & mode 2\\
\cline{1-4}
\multirow{3}{*}{1}
& 75 & 23.241 & 23.407\\
& 85 & 23.404 & 23.320\\
& 95 & 23.539 & 24.463\\
\hline
\multirow{3}{*}{2}
& 75 & 43.089 & 41.257\\
& 85 & 36.189 & 36.599\\
& 95 & 38.426 & 38.316\\
\hline
\multirow{3}{*}{4}
& 75 & 52.266 & 53.773\\
& 85 & 41.273 & 43.012\\
& 95 & 38.626 & 37.239\\
\hline
\multirow{3}{*}{8}
& 75 & 78.752 & 72.742\\
& 85 & 57.214 & 49.458\\
& 95 & 40.130 & 50.978\\
\hline
\end{tabular}
\end{table}

\subsection{Performance of the hierarchical MILP and the hybrid MILP and LP methods}

For the hierarchical MILP and the hybrid MILP and LP methods, we mainly focused on the effect of the region size, the global alignment method, and the DC encoding scheme on the recovery performance, and also the run-time performance.

\begin{figure}[!htb]
\centering
\subfloat[]{\includegraphics[width=\twofigwidth]{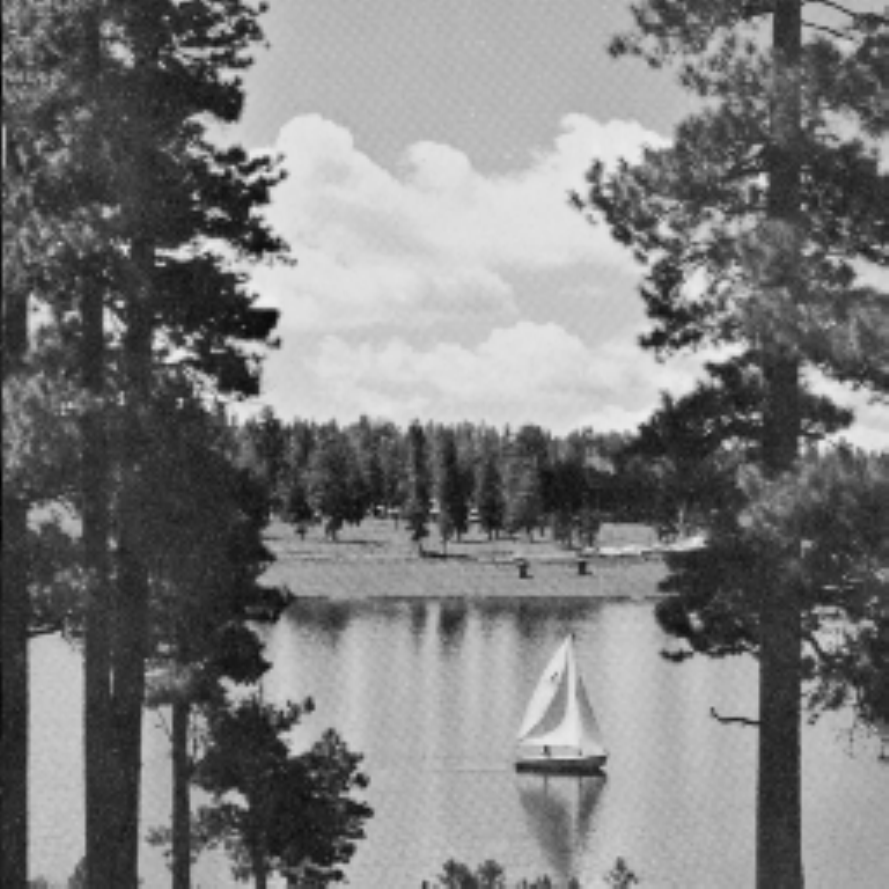}}
\quad
\subfloat[]{\includegraphics[width=\twofigwidth]{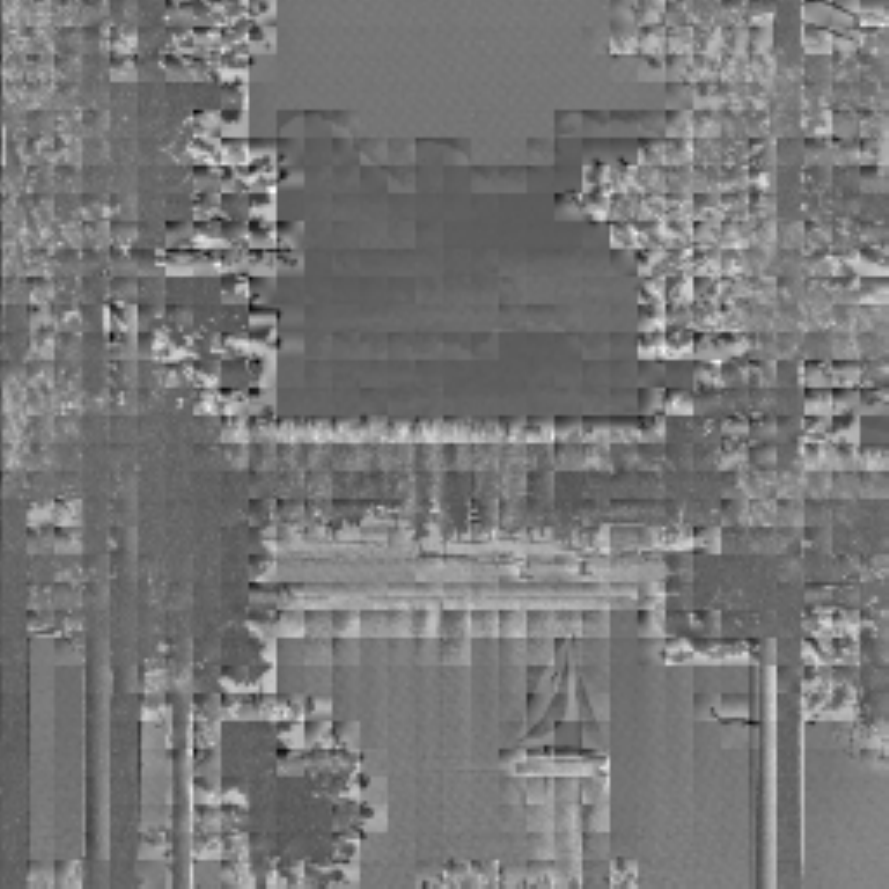}}
\\
\subfloat[]{\includegraphics[width=\twofigwidth]{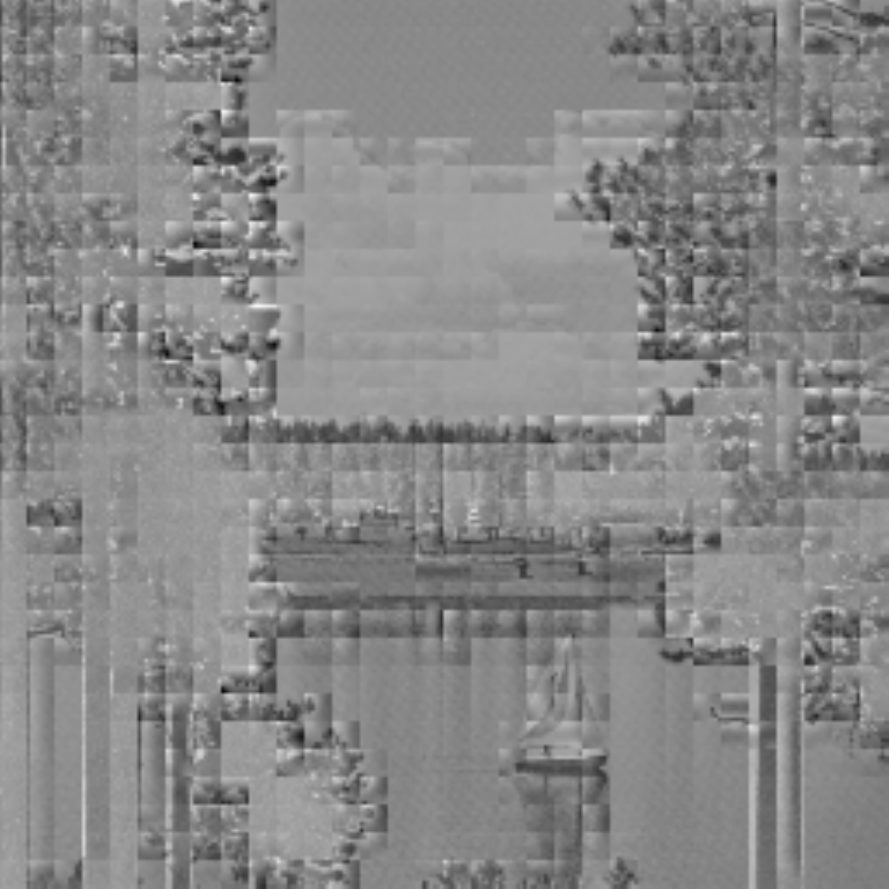}}
\quad
\subfloat[]{\includegraphics[width=\twofigwidth]{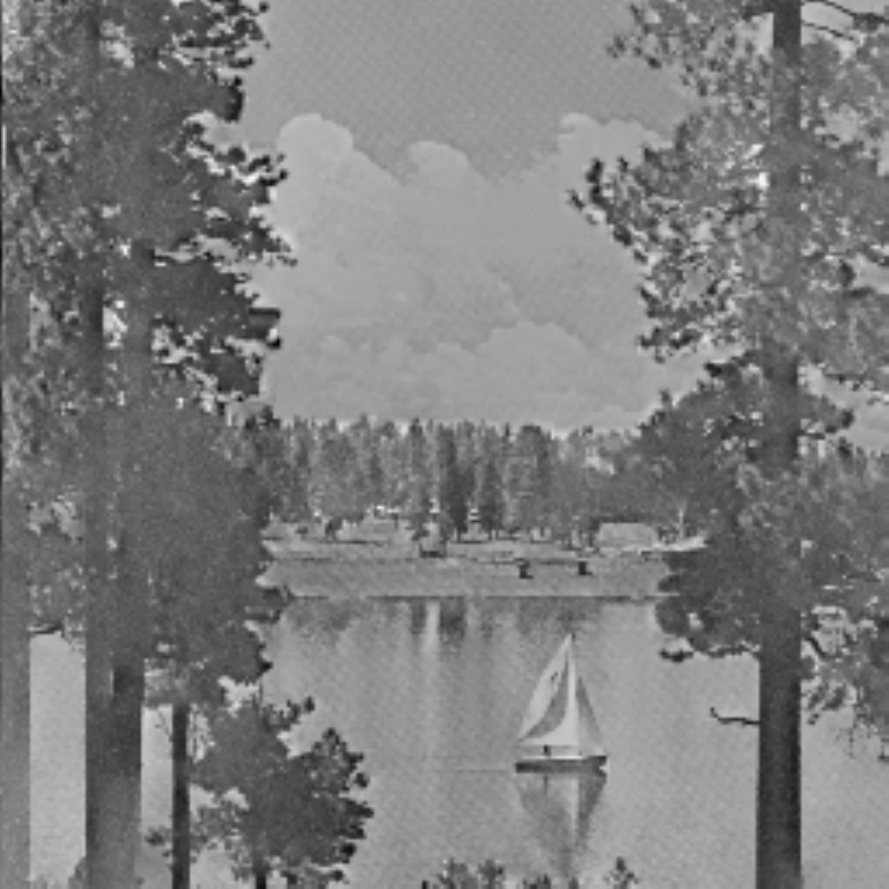}}
\\
\subfloat[]{\includegraphics[width=\twofigwidth]{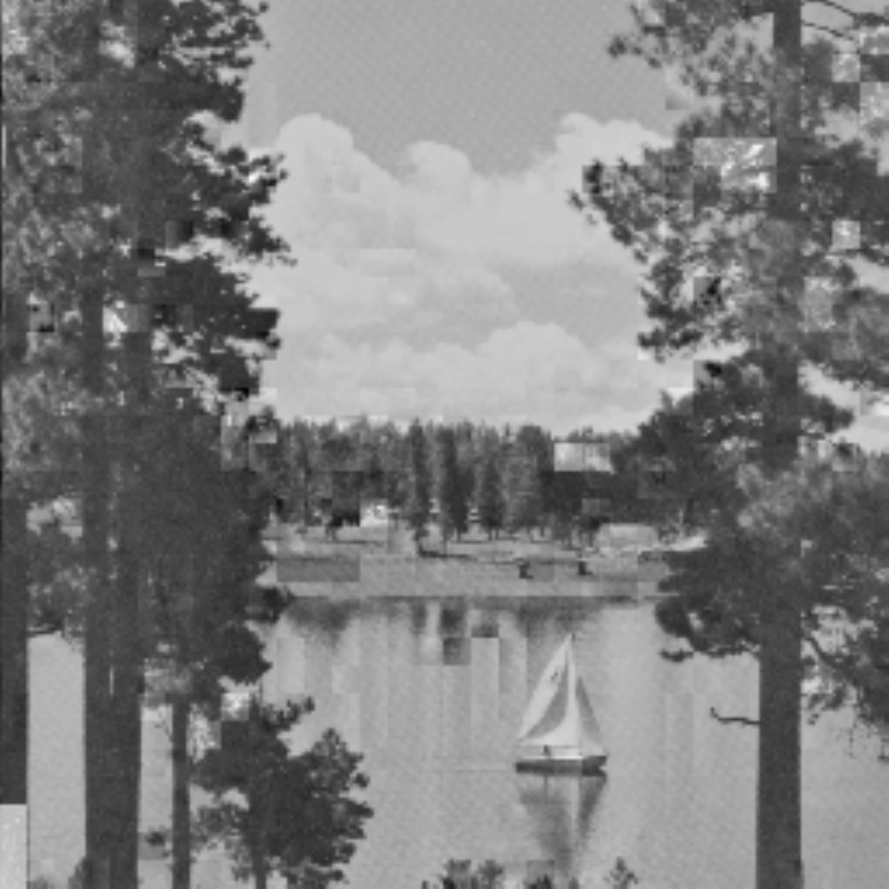}}
\quad
\subfloat[]{\includegraphics[width=\twofigwidth]{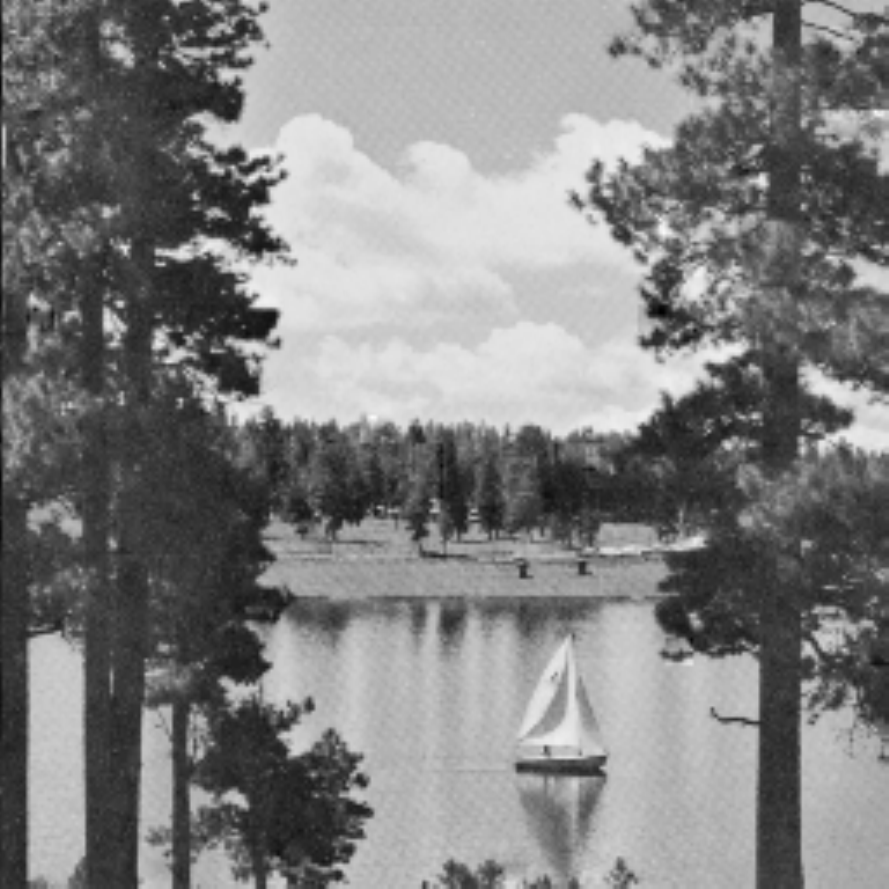}}
\caption{Recovered images with different methods ($U=5$): a) the original image; b) naive (negative); c) naive (positive); d) naive (LP); e) relaxed LP; f) hierarchical MILP.}
\label{fig:comparison}
\end{figure}

\subsubsection{Time complexity}

The time complexity of the regional MILP problems increases exponentially with the number of unknown coefficients. Similar to the relaxed LP method, we attempted to accelerate computation by ignoring small coefficients. However, we found that the acceleration effect varies dramatically across different images. Moreover, even for different regions of the same image, the time consumption can vary by a factor of tens. As a typical example, Table~\ref{table:time-mip} shows the time consumed when solving each of 16 regional MILP problems of a test image of size $256 \times 256$, which is shown in Fig.~\ref{fig:comparison}a), with different thresholds, where region size is $64 \times 64$ and $U=5$. The threshold $T$ does not seem to have a straightforward effect on how the time complexity can be reduced, therefore, we set $T=0$ in subsequent experiments. Besides, we set a time-out value for each MILP problem depending on the problem scale, which was empirically set to be 600 seconds. This time-out value helps assure that the optimization will always finish within a finite time and excludes unrealistic settings that require an excessive amount of time spent for the optimization.

\begin{table}[!htb]
\center
\caption{Time consumed (in seconds) of solving each of 16 regional MILP problems of the test image shown in Fig.~\ref{fig:comparison}a).}
\label{table:time-mip}
\begin{tabular}{*{5}{c|}c}
\hline
\multirow{2}{*}{$T$} & \multirow{2}{*}{row} & \multicolumn{4}{c}{column}\\
\cline{3-6}
& & 1 & 2 & 3 & 4 \\
\hline
\multirow{4}{*}{0}
& 1 & 202.1 & 13.6 & 22.9 & 600\\
& 2 & 123.0 & 18.1 & 33.0 & 498.2\\
& 3 & 231.4 & 241.0 & 278.6 & 493.6\\
& 4 & 38.9 & 14.5 & 232.0 & 12.7\\
\hline
\multirow{4}{*}{5}
& 1 & 171.8 & 5.5 & 4.2 & 600\\
& 2 & 32.9 & 13.5 & 15.5 & 270.1\\
& 3 & 159.5 & 174.6 & 238.6 & 232.7\\
& 4 & 23.4 & 8.5 & 30.3 & 6.2\\
\hline
\multirow{4}{*}{10}
& 1 & 177.5 & 4.2 & 3.9 & 600\\
& 2 & 23.0 & 9.5 & 7.1 & 222.8\\
& 3 & 22.2 & 132.9 & 52.5 & 158.7\\
& 4 & 15.3 & 3.9 & 17.3 & 4.1\\
\hline
\end{tabular}
\end{table}

\subsubsection{How the region size affects the recovery result}

The region size is an essential parameter for the hierarchical MILP and the hybrid MILP and LP methods. In theory, the larger the region size, the better the recovery performance, at the cost of higher time complexity. There should be an adequate large size to balance the recovery performance and the computational complexity. Specifically, increasing the size can only provide a slight improvement in recovery performance, but places an unacceptable burden on computation.

To explore the effect of the region size, we conducted experiments with various region sizes and different values of $U$. Without loss of generality, we divided images into some square regions. The visual quality of recovered images in the first stage is shown in Table~\ref{table:regionsize}. The performance gap between different sizes (from $16 \times 16$ to $64 \times 64$) is consistent across different values of $U$. We observed a big improvement when expanding the size from $16 \times 16$ to $32 \times 32$. Keeping increasing the size, we found that the improvement of the recovery performance was not significant, while the time consumed increased substantially. Although the time limit was already reached when solving the MILP problem for a region of size $64 \times 64$ and $U=7$, we still tried a larger size of $128 \times 128$ with a quadrupled time limit to test the potential of the regional MILP problem-based methods. It was observed that the visual quality instead declined when $U=7$ due to the time restriction. As a whole, $32\times 32$ seems to be a suitable region size, considering both the time complexity and the recovery performance.

\begin{table}[!htb]
\center
\caption{Performance comparison for various region sizes.}
\label{table:regionsize}
\begin{tabular}{*{5}{c|}c}
\hline
\multirow{2}{*}{$U$} & \multirow{2}{*}{Region size} & \multicolumn{2}{c|}{SSIM} & \multicolumn{2}{c}{PSNR}\\
\cline{3-6}
& & mean & median & mean & median\\
\hline
\multirow{4}{*}{3}
& $16 \times 16$ & 0.929371 & 0.930749 & 22.4803 & 22.9342\\
& $32 \times 32$ & 0.974381 & 0.979044 & 31.1263 & 31.0731\\
& $64 \times 64$ & 0.974166 & 0.985446 & 33.7680 & 32.9951\\
& $128\times 128$ & 0.975698 & 0.987922 & 34.0427 & 33.2602\\
\hline
\multirow{4}{*}{5}
& $16 \times 16$ & 0.893620 & 0.895441 & 20.4611 & 20.6191\\
& $32 \times 32$ & 0.961334 & 0.961285 & 28.4843 & 29.2054\\
& $64 \times 64$ & 0.967327 & 0.972655 & 32.0484 & 30.5980\\
& $128\times 128$ & 0.968027 & 0.975557 & 31.9509 & 30.1603\\
\hline
\multirow{4}{*}{7}
& $16 \times 16$ & 0.854239 & 0.849461 & 18.9215 & 19.0734\\
& $32 \times 32$ & 0.946571 & 0.948394 & 27.0813 & 27.0474\\
& $64 \times 64$ & 0.951336 & 0.958771 & 30.4335 & 29.5913\\
& $128\times 128$ & 0.940849 & 0.949296 & 28.8089 & 28.0234\\
\hline
\end{tabular}
\end{table}

\subsubsection{How the global brightness alignment strategy affects the recovery result}

In the second stage, the DC coefficients are further optimized to align the global brightness. We adopted three global alignment strategies including the global MILP, the block LP, and the region LP. The global optimization can be effective for smaller region sizes used in the first stage. In contrast, the additional optimization may be unnecessary when the region size is large enough.

The mean visual quality results of recovered images are shown in Tables~\ref{table:global-ssim} and \ref{table:global-psnr}. Note that we disabled the DC differential encoding here to focus on the alignment effect solely. For comparison, the results in the first stage are also given. There is a noticeable improvement when the region size is $16 \times 16$. Increasing the region size further, the global alignment has negligible effect and even causes a slight decrease in visual quality. Besides, we observed that the global MILP strategies can provide stable and better improvement compared to the other two LP-based alignment strategies. This indicates that the hierarchical MILP method is better than the hybrid MILP and LP methods.

\begin{table}[!htb]
\center
\caption{The mean visual quality of recovered images under various configurations (SSIM).}
\label{table:global-ssim}
\begin{tabular}{c|c*{4}{|c}}
\hline
\multirow{2}{*}{$U$} & \multirow{2}{*}{Region size} & \multirow{2}{*}{First stage} & \multirow{2}{*}{Global MILP} & \multicolumn{2}{c}{LP}\\
\cline{5-6}
& & & & block & region\\
\hline
\multirow{3}{*}{3}
& $16 \times 16$ & 0.929371 & 0.951401 & 0.893120 & 0.903151\\
& $32 \times 32$ & 0.974381 & 0.965687 & 0.915242 & 0.948924\\
& $64 \times 64$ & 0.974166 & 0.965202 & 0.918519 & 0.970792\\
\hline
\multirow{3}{*}{5}
& $16 \times 16$ & 0.893620 & 0.925594 & 0.853869 & 0.859923\\
& $32 \times 32$ & 0.961334 & 0.954448 & 0.900525 & 0.934354\\
& $64 \times 64$ & 0.967327 & 0.958604 & 0.909761 & 0.963775\\
\hline
\multirow{3}{*}{7}
& $16 \times 16$ & 0.854239 & 0.891932 & 0.825837 & 0.826367\\
& $32 \times 32$ & 0.946571 & 0.934435 & 0.881744 & 0.916953\\
& $64 \times 64$ & 0.951336 & 0.941628 & 0.894671 & 0.946944\\
\hline
\end{tabular}
\end{table}

\begin{table}[!htb]
\center
\caption{The mean visual quality of recovered images under various configurations (PSNR).}
\label{table:global-psnr}
\begin{tabular}{c|c*{4}{|c}}
\hline
\multirow{2}{*}{$U$} & \multirow{2}{*}{Region size} & \multirow{2}{*}{First stage} & \multirow{2}{*}{Global MILP} & \multicolumn{2}{c}{LP}\\
\cline{5-6}
 & & & & block & region\\
\hline
\multirow{3}{*}{3}
& $16 \times 16$ & 22.4803 & 28.5883 & 23.4662 & 24.1287\\
& $32 \times 32$ & 31.1263 & 32.0510 & 24.6285 & 29.7521\\
& $64 \times 64$ & 33.7680 & 31.2680 & 25.2650 & 32.2096\\
\hline
\multirow{3}{*}{5}
& $16 \times 16$ & 20.4611 & 27.2269 & 21.6070 & 22.3115\\
& $32 \times 32$ & 28.4843 & 29.7011 & 24.1580 & 28.2956\\
& $64 \times 64$ & 32.0484 & 30.2606 & 25.1237 & 30.8900\\
\hline
\multirow{3}{*}{7}
& $16 \times 16$ & 18.9215 & 25.2569 & 20.5880 & 20.8274\\
& $32 \times 32$ & 27.0813 & 27.9677 & 22.9820 & 26.7920\\
& $64 \times 64$ & 30.4335 & 29.4926 & 24.2791 & 29.6749\\
\hline
\end{tabular}
\end{table}

\begin{figure}[!htb]
\centering
\subfloat[]{\includegraphics[width=\twofigwidth]{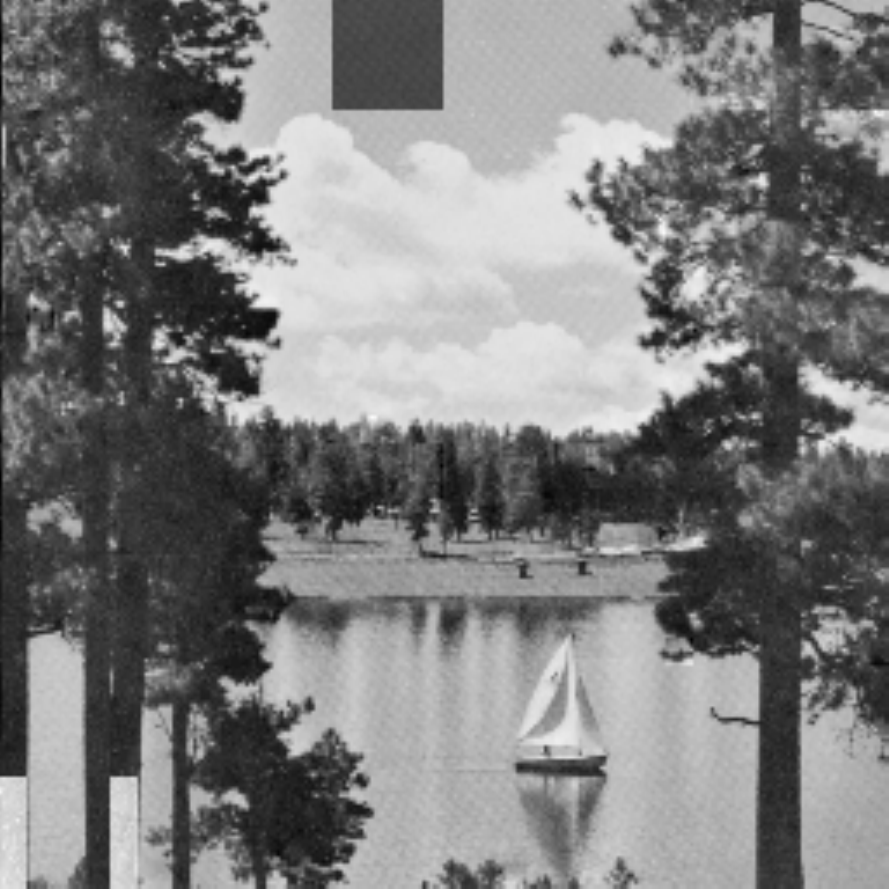}}
\quad
\subfloat[]{\includegraphics[width=\twofigwidth]{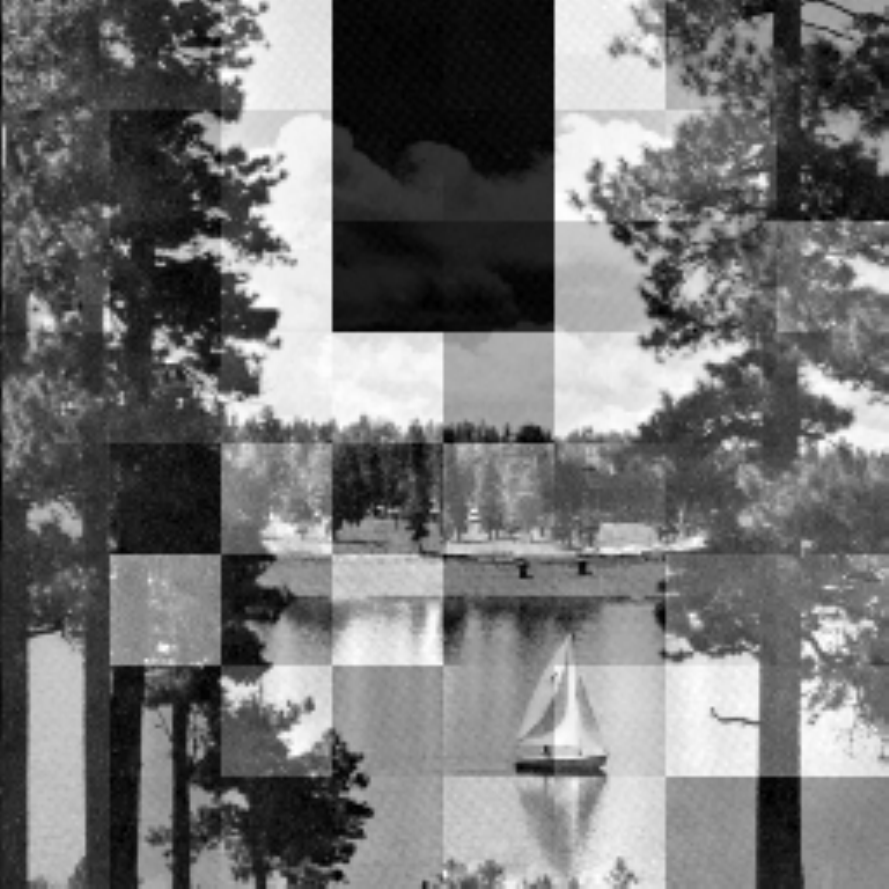}}
\caption{Recovered images using dependency mode 0 in the first stage for different prediction modes: a) mode 0; b) mode 1.}
\label{fig:mip-1st}
\end{figure}

\subsubsection{How the DC encoding scheme and the dependency mode affect the recovery result}

The DC differential encoding causes dependency between divided regions. We designed three DC dependency modes in Sec.~\ref{ssec:ext} to solve the dependency. Here, the global alignment is necessary for dependency mode 0, since the dependency is simply removed, which leads to severe brightness misalignment, as shown in Fig.~\ref{fig:mip-1st}. The global MILP alignment strategy was adopted in the second stage.

Compared with the simple removal of dependency in mode 0, more subtle strategies are developed in the other two modes to resolve the dependency, in the hope of achieving a better performance. However, the error propagation effect should also be considered when the DC differential encoding is considered. Our experimental results showed that the error propagation across different regions cannot be handled well in mode 1, leading to inferior recovery performance compared to mode 0 across different DC prediction modes. We also found that, for mode 2, a slight improvement over mode 1 can be achieved by considering pixel value differences between adjacent regions in the first stage. For mode 0, there is no obvious difference in the recovery performance among the three DC prediction modes. Table~\ref{table:dcdepend} lists the recovery performance under different DC prediction modes and $U=5$ for 22 images of size $256 \times 256$.\footnote{We focused on images of size $256\times 256$ to ease comparison across different dependency modes by limiting the number of different region sizes for dependency modes 1 and 2.} The region size $32\times 32$ was adopted in most cases as this is the size balancing the computational complexity and the recovery performance well. For DC prediction mode 2, considering the special requirements on the region size, we chose three representative values for the region sizes: $8 \times 128$, $16 \times 256$, and $32 \times 256$ for comparison, where the number of pixels of the first one is the same as that of the size $32 \times 32$. It can be seen that the visual quality is considerably degraded when the DC prediction is introduced. For instance, compared with the case of $U=7$ in Tables~\ref{table:global-ssim} and \ref{table:global-psnr}, the recovery performance under DC prediction mode 2 with $U=5$ is even worse than the case of DC prediction mode 0 with $U=7$, although in the former case, there are fewer sign bits missed per block. Such a phenomenon implies that encrypting the sign bits with the differential DC encoding may provide better security than encrypting more sign bits without it.

\begin{table}[!htb]
\center
\caption{Performance comparison for DC dependency modes.}
\label{table:dcdepend}
\begin{tabular}{*{6}{c|}c}
\hline
\multirow{2}{*}{Prediction mode} & \multirow{2}{*}{Dependency mode} &
\multirow{2}{*}{Region size} & \multicolumn{2}{c|}{SSIM} & \multicolumn{2}{c}{PSNR}\\
\cline{4-7}
& & & mean & median & mean & median\\
\hline
\multirow{3}{*}{1}
& 0 & \multirow{3}{*}{$32 \times 32$} & 0.910119 & 0.943161 & 25.6732 & 25.5816\\
& 1 & & 0.893017 & 0.932397 & 24.9979 & 25.2866\\
& 2 & & 0.917904 & 0.946864 & 24.4904 & 24.5651\\
\hline
\multirow{3}{*}[-5ex]{2}
& 0 & $32 \times 32$ & 0.904061 & 0.936648 & 24.2066 & 24.0686\\
\cline{2-2}
& \multirow{3}{*}{1} & $8 \times 128$ & 0.877833 & 0.886896 & 21.5132 & 21.0065\\
& & $16 \times 256$ & 0.927809 & 0.929456 & 24.5208 & 23.9448\\
& & $32 \times 256$ & 0.874819 & 0.905813 & 22.6957 & 21.8742\\
\cline{2-2}
& \multirow{3}{*}{2} & $8 \times 128$ & 0.840171 & 0.866169 & 19.5895 & 19.9287\\
& & $16 \times 256$ & 0.929305 & 0.949870 & 23.9961 & 24.6377\\
& & $32 \times 256$ & 0.884110 & 0.902786 & 22.9986 & 20.5920\\
\hline
\multirow{3}{*}{3}
& 0 & \multirow{3}{*}{$32 \times 32$} & 0.924762 & 0.948330 & 25.5435 & 24.6833\\
& 1 & & 0.892731 & 0.924572 & 23.0082 & 22.1380\\
& 2 & & 0.930015 & 0.957233 & 24.6677 & 23.9453\\
\hline
\end{tabular}
\end{table}

\subsection{Performance comparison}

Two previous subsections show results about some key parameters and methods for identifying suitable settings for our recovery methods. For the hierarchical MILP method, the region size $32 \times 32$ was selected to balance the recovery performance and the computation complexity. To show the effectiveness of our methods, we present three naive methods as benchmarks, since there are no other known solutions based on more advanced techniques in the literature. The naive recovery methods just attempt to restore the unknown parts with simple error-concealment strategies. For instance, the introduced method ``naive (negative)" in Table~\ref{table:comparison} just sets the signs of missing coefficients to negative, while ``naive (positive)" is the converse. For ``naive (LP)", the unknown coefficients are directly assigned by the results obtained in the linear relaxation problem.

By exploiting the summarized properties and conducting an optimization process on the unknown coefficients, the two proposed approximation methods remarkably outperform all the naive methods. Table~\ref{table:comparison} shows the recovery performance of relaxed LP, MILP, and the naive methods in DC prediction mode 0. We also show some restored images using different recovery methods in Fig.~\ref{fig:comparison}. The recovered results of the ``naive (negative)" and ``naive (positive)" methods do not display any detailed visual information. For the ``naive (LP)" method, most of the edge information is lost in the recovered image due to the smoothness maximization of the objective function. Comparing Figs.~\ref{fig:comparison}d) and e), one can see that our relaxed LP method restores the edge information much better by employing the sign extraction function and that the hierarchical MILP method produced the best result.

\begin{table}[!htb]
\center
\caption{Performance comparison for recovery methods.}
\label{table:comparison}
\begin{tabular}{c|c|c|c|c|c}
\hline
\multirow{2}{*}{U} & \multirow{2}{*}{Method} & \multicolumn{2}{c|}{SSIM} & \multicolumn{2}{c}{PSNR}\\
\cline{3-6}
& & mean & median & mean & median\\
\cline{1-6}
\multirow{5}{*}{3}
& Naive (negative) & 0.591160 & 0.586278 & 13.2847 & 13.7387\\
& Naive (positive) & 0.570384 & 0.560025 & 11.9694 & 12.0444\\
& Naive (LP) & 0.876770 & 0.897395 & 19.4387 & 19.0746\\
& Relaxed LP & 0.928892 & 0.933274 & 24.7634 & 24.3539\\
& Hierarchical MILP & \textbf{0.965687} & \textbf{0.981965} & \textbf{32.0510} & \textbf{31.2051}\\
\hline
\multirow{5}{*}{5}
& Naive (negative) & 0.537465 & 0.537715 & 12.7910 & 12.4485\\
& Naive (positive) & 0.518649 & 0.491507 & 11.6627 & 11.6393\\
& Naive (LP) & 0.818258 & 0.835865 & 17.3487 & 17.0201 \\
& Relaxed LP & 0.856900 & 0.858506 & 22.4608 & 22.4850\\
& Hierarchical MILP & \textbf{0.954448} & \textbf{0.964019} & \textbf{29.7011} & \textbf{29.7697}\\
\hline
\multirow{5}{*}{7}
& Naive (negative) & 0.492325 & 0.482679 & 12.3514 & 12.1613\\
& Naive (positive) & 0.474957 & 0.449064 & 11.4660 & 11.3202\\
& Naive (LP) & 0.766469 & 0.787071 & 16.0110 & 15.6254\\
& Relaxed LP & 0.812352 & 0.811006 & 21.1414 & 20.5616\\
& Hierarchical MILP & \textbf{0.934435} & \textbf{0.941955} & \textbf{27.9677} & \textbf{28.4694}\\
\hline
\end{tabular}
\end{table}

\section{Further Discussions, Limitations, and Future Work}

The experiments were conducted with 30 standard test images of two standard image sizes ($256\times 256$ and $384\times 256$). Since these images have been widely used in the image processing field and represent a wide range of natural images, we are confident that our experimental results are generalizable to other natural images. We did not attempt to use a larger image dataset because of the time complexity of our experiments. We call other researchers to validate our results using more and/or larger test images. To facilitate further validation of our work, we released the source code of our proposed methods and the test images we used publicly at the following GitHub repository: \url{https://github.com/ChengqingLi/DCT_SBR}.

It deserves noting that Property~\ref{prop:neighbor}, which is the theoretical foundation of our optimization model, applies to natural images captured using a real image-capturing device only. The Laplacian distribution also holds only statistically across many pixel pairs with a diverse range of pixel values, so for smaller images and some images with a smaller range of pixel values (e.g., drawings and cartoons), the distribution may not be correct or accurate. Therefore, the performance of the optimization model for such less natural images may not be as good as for the natural images we tested in our experiments. Adapting the global model to such less natural or non-natural images will require developing a different statistical model, so more future research is needed.

Although the experiments we conducted were for gray-scale images only, the method can be easily applied to RGB color images or multi-spectral images by treating each channel as an independent gray-scale image. It is possible to explore the cross-channel correlation to potentially improve the performance of the multiple independent channel-specific models. Similarly, our sign bit recovery methods can also be applied to recover digital video files by treating each frame as an independent image, and there are possibilities to explore the temporal (inter-frame) correlation to further improve the visual quality of each recovered video frame and of the whole video. Revising our proposed models and methods for such more complicated images and videos and conducting new experiments will require a substantial amount of new work, so can be a future research direction.

For the hierarchical MILP and the hybrid MILP and LP methods, our experimental results showed that there is a generally optimal region size balancing the recovery performance and the run-time performance. Theoretically speaking, increasing the region size to the maximum (the image size itself) should give the best recovery performance as the model is reduced to the original MILP model, however, this will increase the run-time performance exponentially so does not give the best balance. When only reasonable region sizes are considered, our experimental results showed that there could still be an optimal choice in terms of visual quality. According to the SSIM and PSNR values obtained from our experiments, this optimal region size seems to be between $32\times 32$ and $64\times 64$. We do not expect that this can be easily analyzed theoretically due to the complexity of existing methods for solving MILP problems, but leave such a potential theoretical analysis as future work.

The benchmark methods we used are largely naive error-concealment attacks. This is because no other existing methods have been proposed to recover sign bits beyond such naive attacks. We hope our work will inspire other researchers to propose more advanced sign bit recovery methods, which can then be compared with what we present in this paper.

In addition to the above limitations and future research directions, there are other future research work to improve the performance of our proposed models and methods. For instance, more specialized optimization algorithms may be developed to overcome the general difficulties of solving MILP problems (e.g., the min-cost flow algorithm specially designed for recovering missing DC coefficients in digital images based on the blockwise DCT~\cite{Li:LevelingGrid:ALENEX2012}), more than two layers can be used in the hierarchical MILP method to balance the visual quality and the time complexity, location-varying region sizes depending on the smoothness of different regions can be considered, advanced machine learning methods can be used to help construct more image-specific optimization models, and extend our models and methods to other transforms used in multimedia coding standards (e.g., DWT in JPEG 2000).

\section{Conclusion}

This paper reports our comprehensive study on the sign bit recovery problem of DCT coefficients in digital images. The problem is modeled as a mixed integer linear programming (MILP) problem with the aid of two special properties of natural images. The NP-hard optimization problem was practically addressed by two approximate methods, one based on relaxed LP and the other on hierarchical MILP, both yielding reasonably good recovery results. Special considerations are taken into account to apply the approximation methods to JPEG-encoded images. Extensive experiments were conducted under various conditions to demonstrate the significantly better performance of the proposed methods than other known native methods. The results indicate that the proposed methods can be used in many real-world applications, such as attacking selective encryption schemes based on sign-bit encryption and the development of more efficient image compression and error correction schemes. The optimization models and methods we proposed and experiments we conducted can be further improved and generalized to handle different types of images and videos, so we call for more researchers to conduct follow-up research.

\section*{Acknowledgements}

The work of Chengqing Li was partly supported by the National Natural Science Foundation of China, under the reference numbers 92267102 and 61772447.

\bibliographystyle{elsarticle-num}
\bibliography{main}

\end{document}